\PassOptionsToPackage{unicode}{hyperref}
\PassOptionsToPackage{hyphens}{url}
\documentclass[
]{article}
\usepackage{xcolor}
\usepackage[letterpaper,margin=1in]{geometry}
\usepackage{amsmath,amssymb}
\usepackage{iftex}
\ifPDFTeX
  \usepackage[T1]{fontenc}
  \usepackage[utf8]{inputenc}
  \usepackage{newpxtext,newpxmath}
  \usepackage{textcomp} % provide euro and other symbols
  \DeclareUnicodeCharacter{00B1}{\ensuremath{\pm}}
  \DeclareUnicodeCharacter{00C5}{\AA}
  \DeclareUnicodeCharacter{03B2}{\ensuremath{\beta}}
  \DeclareUnicodeCharacter{03C4}{\ensuremath{\tau}}
  \DeclareUnicodeCharacter{2191}{\ensuremath{\uparrow}}
  \DeclareUnicodeCharacter{2192}{\ensuremath{\to}}
  \DeclareUnicodeCharacter{2193}{\ensuremath{\downarrow}}
  \DeclareUnicodeCharacter{2212}{\ensuremath{-}}
  \DeclareUnicodeCharacter{2248}{\ensuremath{\approx}}
\else % if luatex or xetex
  \usepackage{unicode-math} % this also loads fontspec
  \defaultfontfeatures{Scale=MatchLowercase}
  \defaultfontfeatures[\rmfamily]{Ligatures=TeX,Scale=1}
\fi
\IfFileExists{upquote.sty}{\usepackage{upquote}}{}
\makeatletter
\@ifundefined{KOMAClassName}{% if non-KOMA class
  \IfFileExists{parskip.sty}{%
    \usepackage{parskip}
    \setlength{\parindent}{0pt}
    \setlength{\parskip}{3pt plus 1pt minus 1pt}
  }{% else
    \setlength{\parindent}{0pt}
    \setlength{\parskip}{3pt plus 1pt minus 1pt}}
}{% if KOMA class
  \KOMAoptions{parskip=half}}
\makeatother
\usepackage{color}
\usepackage{fancyvrb}

\DefineVerbatimEnvironment{Highlighting}{Verbatim}{commandchars=\\\{\},fontsize=\small}
\usepackage{longtable,booktabs,array}
\usepackage{multirow}
\usepackage{calc} % for calculating minipage widths
\usepackage{etoolbox}
\AtBeginEnvironment{longtable}{\footnotesize}
\makeatletter
\patchcmd\longtable{\par}{\if@noskipsec\mbox{}\fi\par}{}{}
\patchcmd\LT@output{\vss}{\vskip 0pt plus \maxdimen minus \normalbaselineskip}{}{\PackageError{main}{Could not patch longtable split glue}{}}
\patchcmd\LT@output{\vss}{\vskip 0pt plus \maxdimen minus \normalbaselineskip}{}{\PackageError{main}{Could not patch longtable split glue}{}}
\makeatother
\IfFileExists{footnotehyper.sty}{\usepackage{footnotehyper}}{\usepackage{footnote}}
\makesavenoteenv{longtable}
\usepackage{graphicx}
\usepackage{float}
\usepackage[font=small,labelfont=bf]{caption}
\makeatletter
\newsavebox\pandoc@box
\newcommand*\pandocbounded[1]{% scales image to fit in text height/width
  \sbox\pandoc@box{#1}%
  \Gscale@div\@tempa{\textheight}{\dimexpr\ht\pandoc@box+\dp\pandoc@box\relax}%
  \Gscale@div\@tempb{\linewidth}{\wd\pandoc@box}%
  \ifdim\@tempb\p@<\@tempa\p@\let\@tempa\@tempb\fi% select the smaller of both
  \ifdim\@tempa\p@<\p@\scalebox{\@tempa}{\usebox\pandoc@box}%
  \else\usebox{\pandoc@box}%
  \fi%
}
\def\fps@figure{htbp}
\makeatother
\providecommand{\tightlist}{%
  \setlength{\itemsep}{0pt}\setlength{\parskip}{0pt}}
\usepackage{cite}
\usepackage{bookmark}
\IfFileExists{xurl.sty}{\usepackage{xurl}}{} % add URL line breaks if available
\hypersetup{
  hidelinks,
  pdftitle={Agentic-MatriBO},
  pdfauthor={Yu Zeng, Yuchuan Chen, Ziyue Zhang, Zijian Wei, Yanyi Su, Guojiang Zhao, Hongshuai Wang, Zhifeng Gao},
  pdfcreator={LaTeX}}

\title{Agentic-MatriBO: Capability-Compiled Agentic Bayesian Optimization for Materials and Chemical Design}
\author{\small
\makebox[\textwidth][c]{%
\begin{tabular}{@{}c@{\hspace{5.0em}}c@{\hspace{5.0em}}c@{}}
\begin{tabular}{@{}c@{}}
{\normalsize Yu Zeng\textsuperscript{*}}\\
DP Technology\\
Beijing, China\\
zengyu@dp.tech
\end{tabular}
&
\begin{tabular}{@{}c@{}}
{\normalsize Yuchuan Chen\textsuperscript{*}}\\
DP Technology\\
Beijing, China\\
chenyuchuan@dp.tech
\end{tabular}
&
\begin{tabular}{@{}c@{}}
{\normalsize Ziyue Zhang}\\
DP Technology\\
Beijing, China\\
zhangziyue@dp.tech
\end{tabular}\\[2.8em]
\begin{tabular}{@{}c@{}}
{\normalsize Zijian Wei}\\
DP Technology\\
Beijing, China\\
weizijian@dp.tech
\end{tabular}
&
\begin{tabular}{@{}c@{}}
{\normalsize Yanyi Su}\\
DP Technology\\
Beijing, China\\
suyanyi@dp.tech
\end{tabular}
&
\begin{tabular}{@{}c@{}}
{\normalsize Guojiang Zhao}\\
DP Technology\\
Beijing, China\\
zhaogj@dp.tech
\end{tabular}\\[2.8em]
\begin{tabular}{@{}c@{}}
{\normalsize Hongshuai Wang\textsuperscript{\textdagger}}\\
DP Technology\\
Beijing, China\\
wanghongshuai@dp.tech
\end{tabular}
&
\begin{tabular}{@{}c@{}}
{\normalsize Zhifeng Gao\textsuperscript{\textdagger}}\\
DP Technology\\
Beijing, China\\
gaozf@dp.tech
\end{tabular}
& \mbox{}
\end{tabular}}\\[0.35em]
}
\date{}

\begin{document}
\maketitle
\begingroup
\renewcommand{\thefootnote}{\fnsymbol{footnote}}
\footnotetext[1]{These authors contributed equally to this work.}
\footnotetext[2]{Corresponding authors: Zhifeng Gao (gaozf@dp.tech) and Hongshuai Wang (wanghongshuai@dp.tech).}
\endgroup

\begin{abstract}

Materials and chemical optimization must choose the next synthesis,
characterization, or high-fidelity simulation from sequential
observations under a limited budget. Large language models can use task
knowledge to propose adjustments, but composing surrogates, acquisition
functions, and search strategies is constrained by compatibility,
parameter dependencies, and evidence requirements. We present
\textbf{Agentic-MatriBO}, a capability-compiled agentic Bayesian
optimization (BO) framework that represents BO components and selection
conditions as composable capabilities. A language policy proposes BO
recipes and candidate plans; a compiler checks proposals against the
registered capability catalog; and a deterministic execution layer
validates and commits candidates. Oracle observations are stored in
research memory and inform later rounds. We compare Agentic-MatriBO with
eight baselines on four offline finite candidate pools---COF, PCE10,
Buchwald--Hartwig (BH), and Suzuki--Miyaura---using five seeds per
condition; results are reported descriptively.
Across the four tasks, Agentic-MatriBO has the best mean endpoint quality
after orienting each objective so that higher is better. On COF, it
reaches the tabulated optimum in all five seeds both in the end-to-end
comparison and after shared initialization,
where it also retains an advantage in cumulative quality. Its normalized
area under the best-so-far quality curve during the BO stage (nAUC-BO) is
substantially higher on COF, near parity with the best baseline on PCE10
and BH, and lower on Suzuki. COF ablations show that removing the
capability compiler or evidence gate coincides with lower cumulative
quality and fewer exact-optimum hits. The framework provides an
inspectable interface for evidence-conditioned BO adaptation.

\end{abstract}

\section{1. Introduction}\label{introduction}

In materials and chemical research, a limited budget must be allocated
across candidate syntheses, characterizations, or high-fidelity
simulations. Bayesian optimization (BO) combines a
probabilistic surrogate with an acquisition function to balance
exploitation and exploration in this setting
\cite{lookman2019active,liang2021benchmarking,jones1998ego,shahriari2016review,frazier2018bo}.
Its behavior also depends on candidate representation, initialization,
and configuration, which can be adjusted as evidence accumulates
\cite{liang2021benchmarking,rohr2020benchmarking}.

Numerical BO already adapts acquisition portfolios, local search regions,
and surrogate configurations \cite{hoffman2011portfolio,eriksson2019turbo,lei2021adaptive}.
However, combining heterogeneous components requires explicit
compatibility, evidence, and execution interfaces; the optimizer must
coordinate these components rather than select one rule in isolation.

Large language models (LLMs) provide a natural-language interface for
task semantics and history, but their proposals still require candidate
validity checks and numerical modeling. Existing work injects language
information into representations, kernels, acquisition strategies, or
search regions \cite{cisse2025bora,suwandi2025cake,rankovic2025gollum,ngo2026lmabo,liu2024llambo}.
Turning these capabilities into compositions additionally requires
cross-component contracts and execution rules.

We present Agentic-MatriBO, which organizes two-level compositional
adaptation for materials and chemical optimization through
evidence-conditioned capability compilation. BO components and their
selection conditions are registered as composable capabilities; a
language policy proposes algorithmic adjustments on this basis; and an
execution layer checks and enacts the proposals. Semantic reasoning
proposes the optimization scheme; numerical surrogates and acquisitions
rank candidates, while deterministic checks govern commitment.
Here, the two levels are BO-recipe configuration and per-round candidate
planning.

Specifically, the \textbf{Atomic Capability Space} provides a unified
description of component parameter domains, compatibility relations, and
runtime conditions. The \textbf{Evidence-Conditioned Capability
Compiler} organizes available capabilities from the task and current
state, supports policy selection with historical evidence, and checks
proposal admissibility. \textbf{Two-Level Compositional Adaptation}
jointly proposes BO recipes and candidate plans, while a deterministic
execution layer performs numerical search, candidate validation, and
commitment. The experiment oracle returns target observations, and
research memory feeds observations and reflections into later rounds.
Figure~\ref{fig:overview} shows the overall loop, and Figure~\ref{fig:contracts} expands the capability
contracts, compilation, and execution mechanisms.

The main contributions are:

\begin{itemize}
\tightlist
\item
  \textbf{Capability compilation with selection contracts.} Numerical
  components are registered as atomic capabilities together with their
  parameter domains, compatibility conditions, and evidence
  requirements. The compiler organizes a capability catalog from the
  current search state and checks the admissibility of composition
  proposals.
\item
  \textbf{Execution semantics for two-level adaptation.} The method
  specifies coordinated adjustment of recipes and candidate plans,
  separates policy proposals from candidate commitment, and states the
  conditions under which observation and budget accounting are invariant
  and classical BO is recovered as a special case.
\item
  \textbf{Evaluation that separates the starting point from subsequent
  search.} Four-task comparisons, shared initialization on COF, and
  grouped ablations examine the effects of the starting point and
  configuration, and characterize how optimization gains vary by task
  and search stage.
\end{itemize}

\section{2. Problem Definition}\label{problem-definition}

\subsection{2.1 Standard Bayesian Optimization and Fixed
Policies}\label{standard-bayesian-optimization-and-fixed-policies}

Materials synthesis, characterization, and high-fidelity simulation are
often expensive to evaluate, so a finite budget \(B\) must be used to
search sequentially for high-value regions of a black-box objective
\(f\) \cite{lookman2019active,liang2021benchmarking,jones1998ego,shahriari2016review,frazier2018bo}. Let \(\mathcal H_t\) denote the current completed
input--observation history and let \(\mathcal X_t^{\mathrm{feas}}\) be
the nonempty set of inputs that may be evaluated. Standard BO constructs
a probabilistic surrogate \(p_{m_t}(f\mid\mathcal H_t)\) from the
history, uses an acquisition function to propose a candidate, and
submits it to the evaluation process. In the single-point setting, while
the budget remains and the maximum acquisition value can be obtained,
proposal and commitment can be written as
Eq.~\ref{eq:standard-bo}:

\begin{equation}\label{eq:standard-bo}
\widetilde x_{t+1}\in\arg\max_{x\in\mathcal X_t^{\mathrm{feas}}}
\alpha_t\!\left(x;p_{m_t}(f\mid\mathcal H_t)\right),\qquad
x_{t+1}=\mathsf{Commit}_0(\widetilde x_{t+1}).
\end{equation}

The standard loop commits the proposed candidate and updates the history
only after the target observation is returned. Let
\(m_t,\alpha_t,s_t,\theta_t\) denote the surrogate, acquisition
function, search strategy, and parameters, and let \(\xi_t\) denote
task and runtime state. The fixed policy \(\Pi_0\) maps this interface
to the configuration:

\[
c_t=(m_t,\alpha_t,s_t,\theta_t)=\Pi_0(\mathcal H_t,\xi_t).
\]

Here ``fixed'' refers to the policy and interface, not to an unchanged
configuration: portfolios and search regions may vary with the data
\cite{hoffman2011portfolio,eriksson2019turbo,lei2021adaptive}. We use this
boundary to study contract-governed composition while keeping the task
fixed.

\subsection{2.2 Materials-Task Contracts and a Variable
Optimizer}\label{materials-task-contracts-and-a-variable-optimizer}

A scientific task remains fixed within one optimization campaign and is
represented by the following task tuple:

\[
\tau=(\mathcal I,\mathcal X,x,f_\tau,\mathcal C,B,\nu).
\]

The tuple records candidate identifiers and features, the oracle
objective, feasibility constraint, evaluation budget, and task metadata.
We use maximization notation; minimization tasks use the historical
minimum while retaining the original objective units.

Completed observations and eligible candidates that have not yet been
queried are

\[
\mathcal D_t=\{(i,x_i,y_i):i\in\mathcal I_t^{\mathrm{obs}}\},\qquad
 y_i=f_\tau(i),\qquad
\mathcal I_t^{\mathrm{elig}}=\{i\in\mathcal I\setminus\mathcal I_t^{\mathrm{obs}}:\mathcal C(i)=1\}.
\]

Here \(t\) indexes optimization rounds, and \(n_t=|\mathcal D_t|\)
counts completed evaluations, including initialization. Candidates are
determined by table identifiers, while features are used for numerical
computation. With the configuration notation above, the online optimization
configuration is the recipe

\[
r_t=(m_t,\alpha_t,s_t,\theta_t),
\]

corresponding to the surrogate, acquisition function, search strategy,
and their active parameters. Initialization produces \(\mathcal D_0\) and
an initial recipe, and its oracle calls count toward the budget. Each
round evaluates one candidate while \(n_t<B\) and the pool is nonempty.
Probes and historical retrieval inform the policy without adding target
observations and are excluded from the oracle budget.

\subsection{2.3 Separation of Authority and the Commitment
Boundary}\label{separation-of-authority-and-the-commitment-boundary}

A proposal supplies an optimization adjustment to be checked, while a
commitment authorizes one target evaluation and an observation records
the result returned by the evaluation process. In what follows,
candidate commitment refers to a non-STOP path; STOP terminates the
search and returns a report. The outputs and permissions of the roles
are:

{
\begin{longtable}[]{@{}
  >{\raggedright\arraybackslash}p{(\linewidth - 4\tabcolsep) * \real{0.3333}}
  >{\raggedright\arraybackslash}p{(\linewidth - 4\tabcolsep) * \real{0.3333}}
  >{\raggedright\arraybackslash}p{(\linewidth - 4\tabcolsep) * \real{0.3333}}@{}}
\caption[Roles, permissions, and commitment boundaries]{\textbf{Roles, permissions, and commitment boundaries.}}\label{tab:roles}\\
\toprule\noalign{}
\begin{minipage}[c]{\linewidth}\raggedright
Role
\end{minipage} & \begin{minipage}[c]{\linewidth}\raggedright
Outputs and permissions
\end{minipage} & \begin{minipage}[c]{\linewidth}\raggedright
Boundary
\end{minipage} \\
\midrule\noalign{}
\endfirsthead
\caption[]{\textbf{(continued)}}\\
\toprule\noalign{}
\begin{minipage}[c]{\linewidth}\raggedright
Role
\end{minipage} & \begin{minipage}[c]{\linewidth}\raggedright
Outputs and permissions
\end{minipage} & \begin{minipage}[c]{\linewidth}\raggedright
Boundary
\end{minipage} \\
\midrule\noalign{}
\endhead
\bottomrule\noalign{}
\endlastfoot
Policy reasoning layer & Interprets the task and evidence; proposes
recipes, candidate plans, and probes & Does not directly commit
candidate IDs or generate target observations \\
Deterministic BO execution layer & Compiles the catalog, validates
proposals, generates a candidate shortlist, fuses scores, and commits a
candidate & Validation covers only the actual execution path; it does
not generate target values \\
Advisory reviewers & Return bounded task scores and public justification
for given candidates & Cannot add candidates, commit independently, or veto candidates
unilaterally \\
Experiment oracle & Returns target values and related observations for
submitted IDs & Does not replace target values with language-generated
values \\
Research memory & Stores, links, and retrieves results, hypotheses, and
reflections & Does not authorize execution or treat explanations as
observations \\
\end{longtable}
}

The deterministic execution layer is deterministic given a proposal,
candidate order, scores, runtime state, and numerical environment; the
optimization campaign as a whole still includes LLM calls and random
sampling.

\section{3. Related Work}\label{related-work}

\subsection{3.1 Numerical Modeling and Online Adaptation in Materials
Optimization}\label{numerical-modeling-and-online-adaptation-in-materials-optimization}

BO fits a probabilistic surrogate to observed data and uses an
acquisition function to select the next query \cite{jones1998ego,shahriari2016review,frazier2018bo}, and it is
widely used for sequential experimental design under limited materials
budgets \cite{lookman2019active,liang2021benchmarking}. Materials-optimization benchmarks compare search
behavior under different candidate distributions, objectives, and
initialization conditions \cite{rohr2020benchmarking}; closed-loop experimental studies
focus on connecting recommendations to measurements and on supervising
recommendation quality and the operating process \cite{kusne2020closedloop,harris2025oversight}.

Numerical BO adapts at several levels: acquisition portfolios allocate
opportunities among different selection rules \cite{hoffman2011portfolio}, TuRBO adjusts
the local search region \cite{eriksson2019turbo}, and adaptive-surrogate methods change
the model form \cite{lei2021adaptive}. Libraries such as BoTorch provide an
implementation basis for composing numerical components \cite{balandat2020botorch}. Related
work also discusses calibration and sharpness of probabilistic
predictions \cite{gneiting2007calibration}, as well as the relationship between model
uncertainty and optimization regret \cite{foldager2023uncertainties}, offering different
perspectives for diagnosing and evaluating surrogate models.

\subsection{3.2 Different Entry Points for Language Information into
BO}\label{different-entry-points-for-language-information-into-bo}

Existing LLM--BO methods can be distinguished by where they intervene.
LLAMBO uses natural-language task descriptions and history to support
warm starts, candidate generation, and surrogate-like discrimination
\cite{liu2024llambo}; BORA uses language reasoning to identify potential regions
\cite{cisse2025bora}. At the model level, CAKE evolves GP kernels \cite{suwandi2025cake}, while
GOLLuM jointly learns language-representation adaptation parameters and
GP hyperparameters \cite{rankovic2025gollum}. At the decision level, LMABO adjusts
acquisition strategies \cite{ngo2026lmabo}, and LGBO introduces language
preferences \cite{yuan2026lgbo}. LABO treats LLM predictions as a low-cost
information source to be used alongside expensive evaluations \cite{chen2026labo}.
These methods mainly interface with model representations, candidate
preferences, or optimization configurations.

Recent preprints also explore agentic BO at run time. Brunzema et al.
present Sara, an agent operating through the modular lenz BO framework,
with structured interfaces for reconfiguring numerical components
\cite{brunzema2026sara}; DASH combines numerical diagnostics and language reasoning to adapt
the surrogate model and acquisition strategy separately \cite{zhao2026dash}.

Prior work provides numerical components, language representations, and
candidate-guidance mechanisms. We discuss them at a different design
level, placing them within a capability-compilation framework.
Agentic-MatriBO focuses on the selection conditions, composition
admissibility, and execution semantics of these capabilities, and can
incorporate specific components when the corresponding interfaces and
contracts are satisfied. The scope of the quantitative comparison is
given in Section 5.2.

\section{4. Method: Capability-Compiled
Agentic-MatriBO}\label{method-capability-compiled-agentic-matribo}

\subsection{4.1 Method Overview and Execution
Flow}\label{method-overview-and-execution-flow}

\begin{figure}[H]
\centering
\includegraphics[width=\linewidth,height=0.62\textheight,keepaspectratio]{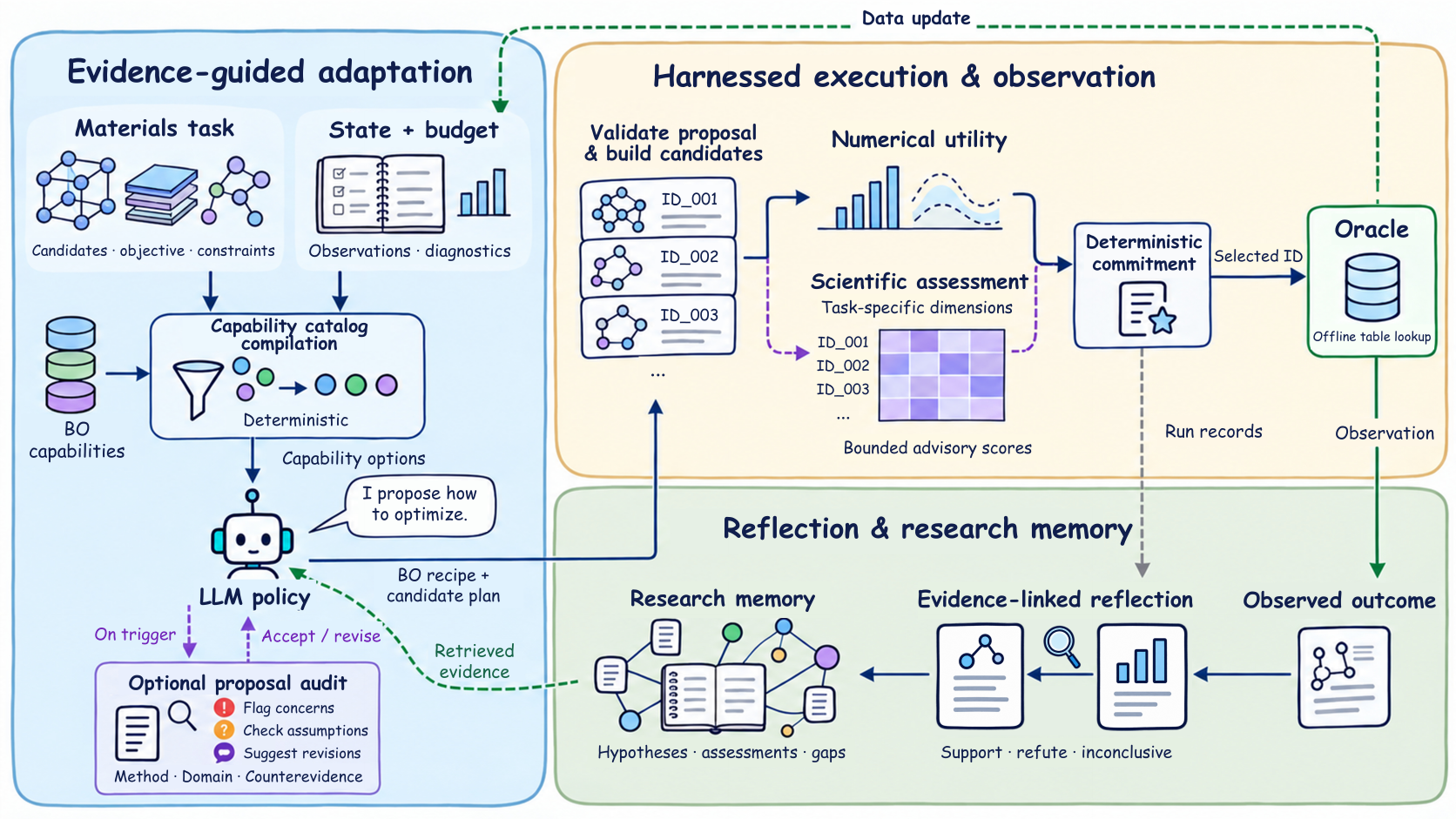}
\caption{\textbf{Research loop and execution boundaries of
Agentic-MatriBO.} The three blocks separate evidence-guided policy
adaptation, candidate decision, and retrospective memory; deterministic
interfaces define their permissions. Capability compilation supplies the
catalog, the policy proposes a recipe and candidate plan, execution
validates and commits a legal candidate, and the oracle returns the
observation. Observations and evidence judgments are stored for later
retrieval, with proposal, candidate, and post-observation reviews applied
at their respective stages.}\label{fig:overview}
\end{figure}

\begin{figure}[H]
\centering
\includegraphics[width=\linewidth,height=0.62\textheight,keepaspectratio]{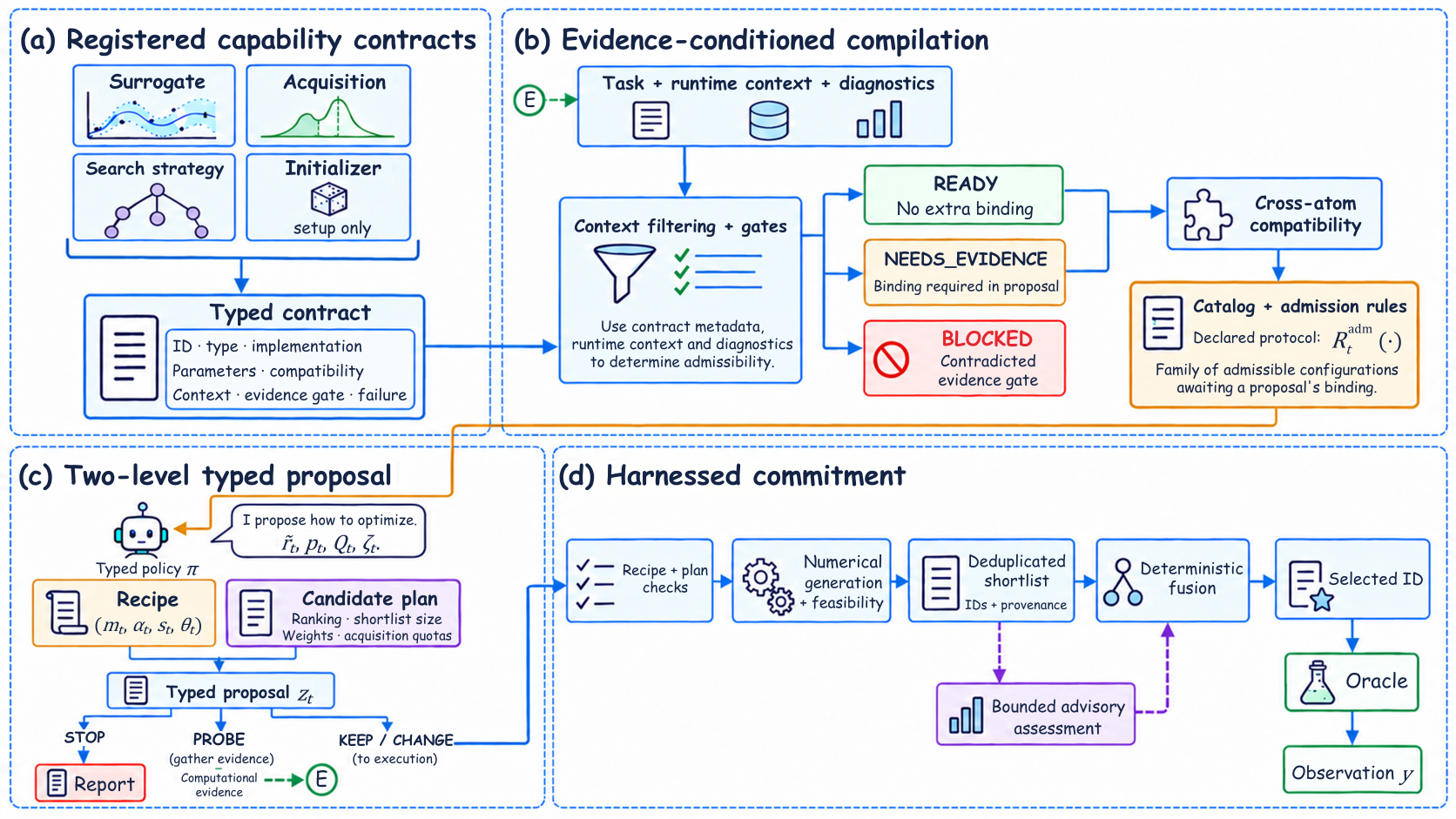}
\caption{\textbf{From capability contracts to candidate commitment.}
(a) Registered capabilities use unified contracts; (b) the compiler
combines task, state, and evidence to define admissible recipes; (c) the
policy proposes a recipe and candidate plan, with PROBE returning
evidence and STOP returning a report; and (d) the execution layer
validates, scores, and commits a legal candidate before the oracle
observation.}\label{fig:contracts}
\end{figure}

Figure~\ref{fig:overview} shows the loop from the search state to a new observation.
\textbf{Evidence-Guided Policy Adaptation} organizes the capability
catalog and evidence to form recipe and candidate-plan proposals;
\textbf{Harnessed Candidate Decision} validates proposals, constructs
the candidate set, and commits a candidate; \textbf{Retrospective
Research Memory} links observations to prior expectations for retrieval
in later rounds.

The recipe-level proposal governs the numerical model and search policy;
the candidate-plan proposal determines how the current round's shortlist
is assembled and ranked. Neither proposal authorizes an oracle call:
deterministic compatibility and feasibility checks precede commitment of
a legal candidate ID, and only the oracle supplies target observations.

\begin{samepage}
\noindent\fbox{%
\begin{minipage}{0.96\linewidth}
\small
\textbf{Algorithm 1. Single-point optimization round under the declared
 target protocol.}\par\smallskip
\textbf{Input:} task \(\tau\), observations \(\mathcal D_t\), recipe \(r_t\),
 runtime \(R_t\), and probe cap \(J_{\max}\).\par
\textbf{1.} If the budget or eligible pool is exhausted, return a report.\par
\textbf{2.} Compile the state, capability catalog, and available evidence.\par
\textbf{3.} The policy proposes an action, recipe, candidate plan, and probes.\par
\textbf{4.} For PROBE, run diagnostics up to \(J_{\max}\) and loop; for STOP,
 return a report.\par
\textbf{5.} Validate contracts, evidence, and the candidate plan; defer or
 repair invalid proposals.\par
\textbf{6.} Build a valid shortlist and fuse deterministic numerical utility
 with complete valid review scores.\par
\textbf{7.} Select a legal candidate ID; if none exists, terminate without
 calling the oracle.\par
\textbf{8.} Evaluate the ID, append the observation, and update memory and
 evidence.\par
\textbf{Output:} the new observation, active recipe, and research state, or
 a termination report.
\end{minipage}}
\end{samepage}

\subsection{4.2 Atomic Capability
Space}\label{atomic-capability-space}

To provide a common interface for heterogeneous adjustments, an
independently selectable capability is represented by the contract
\(a=(\mathrm{id}_a,\kappa_a,\mathsf{Exec}_a,\Theta_a,\mathsf{Comp}_a,
\mathsf{Req}_a,\mathsf{Gate}_a,\mathsf{Fail}_a)\).

The entries are, in order, the canonical ID, capability type, registered
implementation, parameter domain, compatibility conditions, runtime
requirements, evidence prerequisites, and failure signals. The registered
space \(\mathcal A=\mathcal A^m\cup\mathcal A^\alpha\cup\mathcal A^s\cup
\mathcal A^i\) contains surrogate, acquisition, search, and setup-stage
initialization capabilities. An ``atomic'' capability can be selected and
composed independently; its contract exposes applicability, evidence
requirements, and failure signals to the policy and its numerical object
to the execution layer. For example, an ARD GP contract records feature
scale, sample size, and length-scale stability or cross-validation
advantage over an approximately isotropic GP; the compiler turns these
prerequisites into a state that the policy must support.

\subsection{4.3 Evidence-Conditioned Capability
Compiler}\label{evidence-conditioned-capability-compiler}

The catalog-compilation interface \(\Gamma(\tau,S_t,R_t,E_t)\) receives
the task, state, runtime context, and diagnostic or historical evidence.
It filters compatible capabilities, annotates evidence states, and checks
whether proposed compositions and their declared support may enter
execution. Retrieved history informs the policy, whereas catalog states
are computed from encoded task, state, and diagnostic conditions.

For online capability types \(\kappa\in\{m,\alpha,s\}\), the context
catalog is

\[
\mathcal A_t^{\kappa,\mathrm{ctx}}
=\{a\in\mathcal A^\kappa:\mathsf{Comp}_a(\tau)=1,\ \mathsf{Req}_a\subseteq R_t\}.
\]

For each atom in the catalog, task facts, optimization state, and
numerical diagnostics classify its declared minimum evidence
prerequisites as supported, contradicted, or unverified, and reduce them
to

\[
g_t(a)=\begin{cases}
\texttt{BLOCKED},&\text{at least one prerequisite is contradicted},\\
\texttt{READY},&\text{all prerequisites are supported},\\
\texttt{NEEDS\_EVIDENCE},&\text{otherwise}.
\end{cases}
\]

An atom with no minimum evidence prerequisite is READY once its hard
conditions are satisfied; an atom outside the catalog is considered
blocked. The state is determined by encoded admission prerequisites,
while use recommendations and failure signals provide selection guidance
to the policy.

Let \(\zeta_t(a)\) be the list of support items declared for atom \(a\)
by a proposal. Define

\[
b_t(a,\zeta_t)=\mathbf1\{g_t(a)=\texttt{READY}\ \lor\
[g_t(a)=\texttt{NEEDS\_EVIDENCE}\land\zeta_t(a)\ne\varnothing]\}.
\]

The current support-material binding uses a nonempty-list check; it does
not yet verify the existence, visibility, relevance, or degree of
support of a citation for the prerequisite. Let \(K_\tau(r)\) denote the
declared hard contracts for the registered IDs, task and cross-atom
interfaces, and active parameter domains. The strict admissible recipe
space under the target protocol is given by Eq.~\ref{eq:admissible-recipe}:

\begin{align}
\mathcal R_t^{\mathrm{adm}}(\zeta_t)=\{r=(m,\alpha,s,\theta):\;&
 m\in\mathcal A_t^{m,\mathrm{ctx}},\quad \alpha\in\mathcal A_t^{\alpha,\mathrm{ctx}},\nonumber\\[-1pt]
& s\in\mathcal A_t^{s,\mathrm{ctx}},\quad K_\tau(r)=1,\quad b_t(m,\zeta_t)b_t(\alpha,\zeta_t)b_t(s,\zeta_t)=1\}.\label{eq:admissible-recipe}
\end{align}

\subsection{4.4 Two-Level Compositional Adaptation and
Initialization}\label{two-level-compositional-adaptation-and-initialization}

Domain knowledge enters through bounded initialization and online
decisions: initialization selects diverse task-relevant points, while the
policy uses task semantics, diagnostics, budget, and history to propose
configurations and rankings. Reviewers supply advisory scores; target
observations still come only from the oracle.

The policy outputs

\[
z_t=(u_t,\widetilde r_t,p_t,Q_t,\zeta_t),\qquad
u_t\in\{\texttt{KEEP},\texttt{CHANGE},\texttt{PROBE},\texttt{STOP}\}.
\]

The first level proposes a recipe \(\widetilde r_t\); the second proposes
a candidate plan \(p_t=(\mu_t,k_t,\mathbf w_t,\widehat{\mathbf q}_t)\) for
ranking, shortlist size, source weights, and acquisition quotas. Sources
are acquisition, max-variance, farthest, and random-coverage; the policy
also proposes review dimensions and diagnostic requests under a separate
retrieval budget.

The target protocol requires an admissible recipe, compatible acquisition
capabilities with satisfied evidence gates, and legal plan size, weights,
and deduplication. In execution, an evidence gate or cooldown may defer
CHANGE to KEEP; illegal IDs, pairings, and other contract violations
enter repair or failure handling.

PROBE updates computational evidence without updating observations:

\[
E_t^{(j+1)}=E_t^{(j)}\cup\operatorname{Probe}(Q_t),\qquad
\mathcal D_t^{(j+1)}=\mathcal D_t^{(j)}.
\]

A structured controller limits diagnostic cycles to \(J_{\max}\) per
round; further requests become KEEP after the limit. STOP returns a
report, while KEEP and CHANGE proceed to recipe and candidate checks.

Diagnostics can compare surrogates, predictive intervals, ranking
consistency, or candidate-pool posteriors; retrieval supplies earlier
failures and contradictory evidence for the next decision.

An optional proposal audit can recommend revisions from methodological,
domain, or falsification perspectives before execution checks the draft;
its triggers and implementation are given in Appendix C.3.4.

\subsection{4.5 Multi-Source Candidates and Deterministic
Fusion}\label{multi-source-candidates-and-deterministic-fusion}

Candidate composition separates source-slot allocation from
candidate-utility calculation. Let
\(P_t^+=\{\ell:\widehat q_{\ell,t}>0\}\) be the set of
acquisition-function indices with positive requested quotas. When
\(P_t^+\ne\varnothing\), the utility of the acquisition portfolio is

\[
\bar\alpha_t(i)=\sum_{\ell\in P_t^+}
\frac{\widehat q_{\ell,t}}{\sum_{k\in P_t^+}\widehat q_{k,t}}
\rho_{\mathrm{pool}}\bigl(\alpha_{\ell,t}(i)\bigr).
\]

When no positive quota is present, the active acquisition function is
used. \(\rho_{\mathrm{pool}}\) assigns stabilized linear ranks within
the scoring pool, which may be smaller than the full legal pool.

The source weights \(\mathbf w_t=(\omega_a,\omega_\sigma,\omega_d,\omega_r)\)
control both source-slot allocation and mixed numerical utility. The
requested size is clipped to the legal scoring pool and the execution
layer converts source and acquisition quotas into integer slots.

Acquisition, max-variance, farthest, and random-coverage proposals are
merged, deduplicated, and filled according to the execution rules.
Source quotas determine composition, whereas review-dimension weights
determine advisory-score fusion; detailed shortfall and coverage rules
are given in Appendix C.3.

Let \(a_i,v_i,c_i\) be the tolerance-average ranks of acquisition
utility, uncertainty, and coverage distance within the candidate
shortlist. Numerical utility is respectively \(a_i,v_i,c_i\) in exploit,
uncertainty, and coverage modes; in mixed mode, define the numerical
utility \(\nu_i\) by Eq.~\ref{eq:numerical-utility}:

\begin{equation}\label{eq:numerical-utility}
\nu_i=\frac{\omega_a a_i+\omega_\sigma v_i+(\omega_d+\omega_r)c_i}
{\omega_a+\omega_\sigma+\omega_d+\omega_r},\qquad
\omega_a+\omega_\sigma+\omega_d+\omega_r>0.
\end{equation}

Let \(\mathcal V_t\) be the valid review dimensions covering the
shortlist, \(r_{id}\in[0,1]\) their scores, and
\(\widetilde w_{d,t}\) the renormalized weights. With effective
numerical-utility weight \(\lambda_t^{\mathrm{eff}}\), the final score is
Eq.~\ref{eq:fused-score}:

\begin{equation}\label{eq:fused-score}
F_i=\lambda_t^{\mathrm{eff}}\nu_i+
(1-\lambda_t^{\mathrm{eff}})\sum_{d\in\mathcal V_t}\widetilde w_{d,t}r_{id}.
\end{equation}

Review dimensions and weights are proposed by the policy and informed by
configuration, diagnostics, and memory. If no complete valid review set
has positive weight, \(\lambda_t^{\mathrm{eff}}=1\) and selection is
purely numerical. Advisory scores are not probability-calibrated; review
requirements are given in Appendix C.3.4.

A deterministic feasibility check excludes illegal candidates, while a
veto request from an LLM reviewer is treated as advice. Among the legal
candidate shortlist, select according to Eq.~\ref{eq:commitment-update}:

\begin{equation}\label{eq:commitment-update}
i_{t+1}\in\arg\max_i F_i,\qquad
y_{t+1}=f_\tau(i_{t+1}),\qquad
\mathcal D_{t+1}=\mathcal D_t\cup\{(i_{t+1},x_{i_{t+1}},y_{t+1})\}.
\end{equation}

Ties follow the specified deterministic order; after commitment, the
oracle returns the target observation.

\subsection{4.6 Research Hypotheses, Public Reflection, and Memory
Feedback}\label{research-hypotheses-public-reflection-and-memory-feedback}

After each oracle result, the system links the proposal, observation, and
prior expectation, records a SUPPORT, REFUTE, or INCONCLUSIVE judgment,
and feeds hypotheses, evidence gaps, and retrieval rules into later
rounds (Figure~\ref{fig:overview}). Public hypotheses include testable predictions and
falsification conditions; hypotheses about domain effects or interactions
also specify comparison arms, controls, and minimum evidence. Judgments
are descriptive and stored with their basis; detailed states, citation
checks, and update rules are given in Appendix C.3.

\subsection{4.7 Protocol Properties and Conditions for Their
Validity}\label{protocol-properties-and-conditions-for-their-validity}

Properties 1--3 are conditional invariants of the complete protocol;
implementation coverage is summarized in Section 4.8.

\textbf{Property 1: Validity of declared admission.} If registered
implementations agree with their contracts, compatibility predicates are
correct, active parameters are processed within their declared domains,
and the execution layer enforces the strict protocol, then by
construction of the admissible set,
\(r\in\mathcal R_t^{\mathrm{adm}}(\zeta_t)\) satisfies the
registered-ID, context, cross-atom, and required nonempty-binding
conditions.

\textbf{Property 2: Observation and budget accounting on non-STOP
paths.} If initialization and subsequent calls are managed by the same
budget counter, every completed evaluation corresponds to one newly
added candidate, and records have no loss or duplication, then \(|\mathcal D_T|=N_{\mathrm{oracle}}(T)\le B\).

Recipe adjustments and probes do not add targets; only a budget-checked
Experiment transition adds an observation.

\textbf{Property 3: Classical one-step BO as a special case.} If the
recipe is fixed, the full legal pool is scored, a single acquisition
function performs global candidate selection, its maximizer enters the
shortlist and is not subsequently removed, the exploit numerical mode is
used with review influence disabled, the rank transform preserves the
set of acquisition maximizers, and ties are handled consistently, then
\(i_{t+1}\in\arg\max_{i\in\mathcal I_t^{\mathrm{elig}}}\alpha(i;\mathcal D_t)\).

Under these conditions fusion preserves the acquisition ordering; a
bounded scoring pool or tolerance-based ranks can otherwise change the
selected point.

\subsection{4.8 Checking Coverage of the Current
Implementation}\label{checking-coverage-of-the-current-implementation}

The current implementation mainly checks newly switched components on
the CHANGE path. Supporting-material binding, delayed-action
revalidation, and complete ID--feature consistency checks remain
incomplete; the full comparison is provided in Appendix C.4.

\section{5. Experiments}\label{experiments}

Across four finite pools, we compare endpoint and cumulative search
quality, examine initialization and configuration effects through grouped
COF ablations, and test LLM-based selection; a zero-observation audit
tests sensitivity to candidate IDs. Each condition used five seeds; the
strategy-event timelines and search-path projections are provided in
Appendices C.1 and D.2.

\subsection{5.1 Tasks, budget, and evidence
scope}\label{tasks-budget-and-evidence-scope}

{
\begin{longtable}[]{@{}
  >{\raggedright\arraybackslash}p{(\linewidth - 6\tabcolsep) * \real{0.2500}}
  >{\raggedright\arraybackslash}p{(\linewidth - 6\tabcolsep) * \real{0.2500}}
  >{\raggedright\arraybackslash}p{(\linewidth - 6\tabcolsep) * \real{0.2500}}
  >{\raggedright\arraybackslash}p{(\linewidth - 6\tabcolsep) * \real{0.2500}}@{}}
\caption[Tasks, candidate pools, and observation sources]{\textbf{Tasks, candidate pools, and observation sources.}}\label{tab:tasks}\\
\toprule\noalign{}
\begin{minipage}[c]{\linewidth}\raggedright
Task
\end{minipage} & \begin{minipage}[c]{\linewidth}\raggedright
Input space
\end{minipage} & \begin{minipage}[c]{\linewidth}\raggedright
Candidate pool
\end{minipage} & \begin{minipage}[c]{\linewidth}\raggedright
Objective / source
\end{minipage} \\
\midrule\noalign{}
\endfirsthead
\caption[]{\textbf{(continued)}}\\
\toprule\noalign{}
\begin{minipage}[c]{\linewidth}\raggedright
Task
\end{minipage} & \begin{minipage}[c]{\linewidth}\raggedright
Input space
\end{minipage} & \begin{minipage}[c]{\linewidth}\raggedright
Candidate pool
\end{minipage} & \begin{minipage}[c]{\linewidth}\raggedright
Objective / source
\end{minipage} \\
\midrule\noalign{}
\endhead
\bottomrule\noalign{}
\endlastfoot
COF (Xe/Kr) & 14 numerical descriptors & 605 candidates & Maximize
adsorption selectivity; LABO finite table with GCMC targets
\cite{chen2026labo,gantzler2023cof} \\
PCE10 & 4 composition variables & 1,040 records (1,020 compositions) &
Minimize photodegradation; LABO finite table \\
Buchwald--Hartwig & 4 categorical variables & 790 candidates & Maximize
yield; Buchwald--sub4 table \cite{ahneman2018reaction} \\
Suzuki--Miyaura & 5 categorical variables & 5,760 candidates & Maximize
yield; reaction-condition table \cite{perera2018platform} \\
\end{longtable}
}

\textbf{Note.} Each condition comprised five seed-specific campaigns, each with \textbf{10 initialization and 30 subsequent single-point
evaluations} (40 total); observations were drawn from the offline tables.
Each PCE10 record contains four component mass fractions that sum to 1,
leaving three independent degrees of freedom. The 1,040 records represent
1,020 distinct compositions; repeated compositions remain separate
candidate rows with their own row IDs and oracle values. All runs used the
complete candidate pools in their respective tables; data sources and
processing are described in Appendices B.2.3 and B.2.6.

\subsection{5.2 Compared methods and grouped
controls}\label{compared-methods-and-grouped-controls}

The end-to-end comparison on all four tasks included Vanilla BO
\cite{jones1998ego,shahriari2016review,frazier2018bo}, HEBO \cite{cowenrivers2022hebo}, LABO \cite{chen2026labo}, CAKE \cite{suwandi2025cake}, LMABO \cite{ngo2026lmabo},
BORA \cite{cisse2025bora}, Reasoning BO \cite{yang2025reasoningbo}, and LLAMBO \cite{liu2024llambo}. Each method
used its own initialization to determine the initial IDs and then
carried out subsequent search. COF also included a shared-initialization
group: for a given seed, all methods received the same 10 initial
candidates and observations, followed by 30 selections per method.

All baselines operated on the finite pools; the HEBO and Reasoning BO
adapters and fallback details are given in Appendix B.2.5.

COF ablations were divided into three groups. The warm-start group
compared agentic and random initialization. The component group, under a
shared random initialization, compared the complete system, No review,
and fixed GP+UCB. The Harness group, meaning the execution-layer control
that validates, scores, and commits candidates, used another frozen shared
initialization and compared the complete system with four execution-related
variants. The complete-system nAUC-BO values in the three groups were 0.970,
0.888, and 0.923, respectively; each variant was compared within its
own group. The 0.970 and 0.923 values use different initializations, and
the repeated endpoint in the warm-start and component groups is not
treated as an independent replicate.

These groups partially isolate starting-point and configuration effects within their respective controls. Bare LLM is a system-level
control because it retains natural-language interaction but uses a
different action space and failure handling; LLM commitment changes final
candidate selection. In the model-sensitivity experiment, the LLM
participated in initialization and subsequent decisions.

\subsection{5.3 Metrics and statistical
interpretation}\label{metrics-and-statistical-interpretation}

Let \(I_n\) denote the best observed target after the \(n\)th evaluation:
the historical maximum for maximization and minimum for minimization.
Endpoint metrics retain their original units. Let \(y_\star\) and
\(y_{\mathrm{worst}}\) be the best and worst values in the complete table,
fixed before search. Here \(Q_n\) is normalized quality, distinct from
the probe request \(Q_t\) in Section 4.4; the tabulated optimum has
\(Q_n=1\). We use the following fixed-direction normalization:

\begin{equation}\label{eq:metrics}
Q_n=\begin{cases}
\dfrac{I_n-y_{\mathrm{worst}}}{y_\star-y_{\mathrm{worst}}},&\text{maximization},\\
\dfrac{y_{\mathrm{worst}}-I_n}{y_{\mathrm{worst}}-y_\star},&\text{minimization}.
\end{cases}
\end{equation}

The endpoint factor is \(\mathrm{EF40}=Q_{40}\), and the two discrete
trapezoidal areas are

\begin{equation}\label{eq:auc-metrics}
\mathrm{nAUC}_{\mathrm{BO}}=\dfrac1{30}\sum_{n=10}^{39}\dfrac{Q_n+Q_{n+1}}{2},\qquad
\mathrm{nAUC}_{\mathrm{late}}=\dfrac1{20}\sum_{n=20}^{39}\dfrac{Q_n+Q_{n+1}}{2}.
\end{equation}

Normalized metrics are compared only within the same task; no aggregate
cross-task score is computed.

The BO-stage area includes quality at the end of initialization \(I_{10}\);
EF40 and Best-so-far share an endpoint, and late-stage area is contained
in BO-stage area. The COF ablation reports \(t_{95}\), the first
oracle-evaluation index (including initialization) with \(Q_n\ge0.95\);
a non-hit within \(B=40\) is assigned 41. We report its five-seed mean
and threshold-hit count separately from exact-optimum hits.

Unless otherwise stated, means ± s.d. are over five campaign seeds.
Comparisons are descriptive and include no significance, equivalence, or
non-inferiority tests; the primary end-to-end summaries are not uniformly
paired across methods at the per-seed trajectory level. Section 5.10 is a
separate paired original-ID versus anonymous-ID audit, so its McNemar and
sign tests address a different within-trial intervention. Underlying
figure and table values are listed in Appendices A.1--A.3.

\subsection{5.4 Information access and computational
budget}\label{information-access-and-computational-budget}

Information access is restricted to queried targets and evidence
available before the current round. We compare search quality at the
same oracle-call budget; LLM calls, probes, retries, and model fitting
add overhead but are not quantitatively compared here
\cite{eggensperger2021hpobench,bouthillier2021variance,dodge2019show,pineau2021reproducibility}.

\subsection{5.5 Cross-task endpoint performance and cumulative search
quality}\label{cross-task-endpoint-performance-and-cumulative-search-quality}

\begin{figure}[H]
\centering
\includegraphics[width=\linewidth,height=0.62\textheight,keepaspectratio]{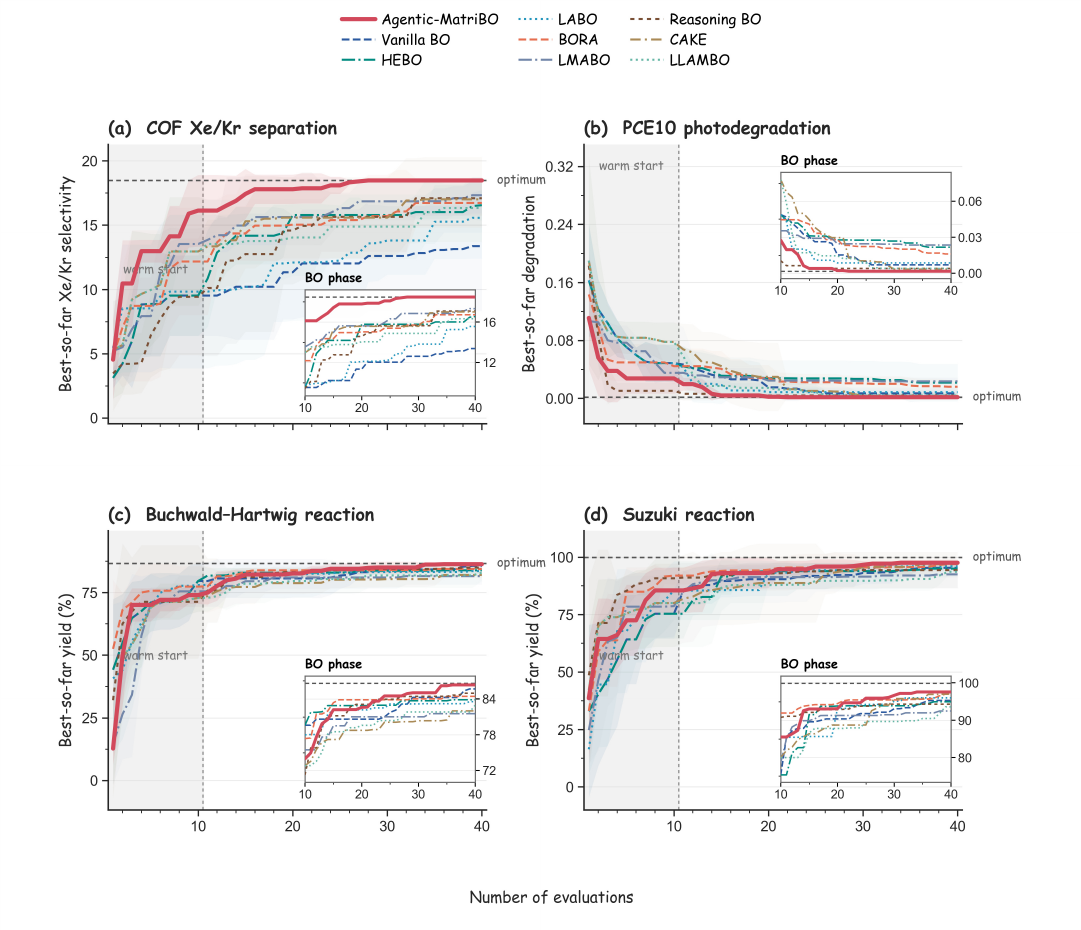}
\caption{\textbf{End-to-end optimization performance on four
tasks.} (a) COF; (b) PCE10; (c) BH; (d) Suzuki. Curves and shaded
regions show the five-seed mean incumbent and ±1 s.d.; the horizontal
axis includes 10 initial oracle evaluations plus 30 subsequent evaluations. Lower is
better for PCE10, whereas higher is better for the other tasks. The gray
region denotes initialization, the inset enlarges the BO stage, and the
dashed line marks the tabulated optimum. Each method uses its own
initialization procedure; the figure compares the complete running
configurations.}\label{fig:end-to-end}
\end{figure}

Across the five-seed summaries, Agentic-MatriBO has the best endpoint
mean on all four tasks among the listed baselines, whereas cumulative
area is not uniformly best (Fig.~\ref{fig:end-to-end} and Table~\ref{tab:endpoints}). Endpoint differences
are larger for COF and PCE10 and smaller for BH and Suzuki.

{
\footnotesize
\begin{longtable}[]{@{}
  >{\raggedright\arraybackslash}p{(\linewidth - 8\tabcolsep) * \real{0.1000}}
  >{\raggedright\arraybackslash}p{(\linewidth - 8\tabcolsep) * \real{0.1900}}
  >{\raggedright\arraybackslash}p{(\linewidth - 8\tabcolsep) * \real{0.1700}}
  >{\centering\arraybackslash}p{(\linewidth - 8\tabcolsep) * \real{0.2600}}
  >{\centering\arraybackslash}p{(\linewidth - 8\tabcolsep) * \real{0.2800}}@{}}
\caption[Endpoint and cumulative metrics across four tasks]{\textbf{Endpoint and cumulative metrics across four tasks.} Values are mean ± s.d. over five seeds. For each task--metric pair, the table reports the best baseline and Agentic-MatriBO. Boldface indicates better Agentic-MatriBO mean performance; a dagger indicates that the two displayed means round to the same value and is not a significance claim.}\label{tab:endpoints}\\
\toprule\noalign{}
\begin{minipage}[c]{\linewidth}\raggedright
Task
\end{minipage} & \begin{minipage}[c]{\linewidth}\raggedright
Metric
\end{minipage} & \begin{minipage}[c]{\linewidth}\raggedright
Best baseline
\end{minipage} & \begin{minipage}[c]{\linewidth}\centering
Baseline score
\end{minipage} & \begin{minipage}[c]{\linewidth}\centering
Agentic-MatriBO
\end{minipage} \\
\midrule\noalign{}
\endfirsthead
\caption[]{\textbf{(continued)}}\\
\toprule\noalign{}
\begin{minipage}[c]{\linewidth}\raggedright
Task
\end{minipage} & \begin{minipage}[c]{\linewidth}\raggedright
Metric
\end{minipage} & \begin{minipage}[c]{\linewidth}\raggedright
Best baseline
\end{minipage} & \begin{minipage}[c]{\linewidth}\centering
Baseline score
\end{minipage} & \begin{minipage}[c]{\linewidth}\centering
Agentic-MatriBO
\end{minipage} \\
\midrule\noalign{}
\endhead
\bottomrule\noalign{}
\endlastfoot
\multirow[t]{4}{*}{\textbf{COF}} & Best-so-far ↑ & LMABO & \mbox{17.326 ± 1.416} & \textbf{\mbox{18.476 ± 0.000}} \\
& nAUC-BO ↑ & LMABO & \mbox{0.868 ± 0.076} & \textbf{\mbox{0.970 ± 0.039}} \\
& EF40 ↑ & LMABO & \mbox{0.938 ± 0.077} & \textbf{\mbox{1.000 ± 0.000}} \\
& Late nAUC ↑ & LMABO & \mbox{0.898 ± 0.075} & \textbf{\mbox{0.991 ± 0.015}} \\
\addlinespace[3pt]
\multirow[t]{4}{*}{\textbf{PCE10}} & Best-so-far ↓ & CAKE & \mbox{0.0034 ± 0.0040} & \textbf{\mbox{0.00162 ± 0.00000}} \\
& nAUC-BO ↑ & Reasoning BO & \mbox{0.9963 ± 0.0062} & \textbf{\mbox{0.9964 ± 0.0021}} \\
& EF40 ↑ & CAKE & \mbox{0.998 ± 0.005} & \textbf{\mbox{1.000 ± 0.000}} \\
& Late nAUC ↑ & Reasoning BO & \mbox{0.997 ± 0.005} & \textbf{\mbox{1.000 ± 0.000}} \\
\addlinespace[3pt]
\multirow[t]{4}{*}{\textbf{BH}} & Best-so-far ↑ & Vanilla BO & \mbox{85.644 ± 1.000} & \textbf{\mbox{86.316 ± 0.385}} \\
& nAUC-BO ↑ & BORA & \mbox{0.963 ± 0.008} & \mbox{0.963 ± 0.011}\(\dagger\) \\
& EF40 ↑ & Vanilla BO & \mbox{0.989 ± 0.012} & \textbf{\mbox{0.997 ± 0.004}} \\
& Late nAUC ↑ & Reasoning BO & \mbox{0.973 ± 0.015} & \textbf{\mbox{0.981 ± 0.015}} \\
\addlinespace[3pt]
\multirow[t]{4}{*}{\textbf{Suzuki}} & Best-so-far ↑ & CAKE & \mbox{96.990 ± 1.580} & \textbf{\mbox{97.558 ± 1.038}} \\
& nAUC-BO ↑ & BORA & \mbox{0.950 ± 0.028} & \mbox{0.946 ± 0.033} \\
& EF40 ↑ & CAKE & \mbox{0.971 ± 0.016} & \textbf{\mbox{0.977 ± 0.010}} \\
& Late nAUC ↑ & BORA & \mbox{0.958 ± 0.030} & \textbf{\mbox{0.965 ± 0.020}} \\
\end{longtable}
}

By task, the method has a better objective-direction-adjusted mean on
all four metrics for COF and PCE10, on three of four metrics for BH (the
nAUC-BO means round to the same value), and on three of four metrics for
Suzuki (nAUC-BO is lower). These comparisons are descriptive rather than a vote across
independent outcomes: Best-so-far and EF40, and late-stage and BO-stage
areas, are derived from the same trajectories (Section 5.3).

On COF, the complete condition reaches the tabulated optimum in all five
seeds (18.476 ± 0.000; 5/5 exact-optimum hits), while LMABO has a lower
mean endpoint. The endpoint is approximately 6.6\% higher than LMABO,
so this difference is primarily an exact-hit comparison. PCE10 shows
a large endpoint difference but only a 0.0001 BO-stage-area difference,
whereas Suzuki has higher endpoint and late-stage area but lower BO-stage
area than BORA. These examples show that endpoint and cumulative quality
can separate by task and stage.

\subsection{5.6 COF search performance after shared
initialization}\label{cof-search-performance-after-shared-initialization}

\begin{figure}[H]
\centering
\includegraphics[width=\linewidth,height=0.40\textheight,keepaspectratio]{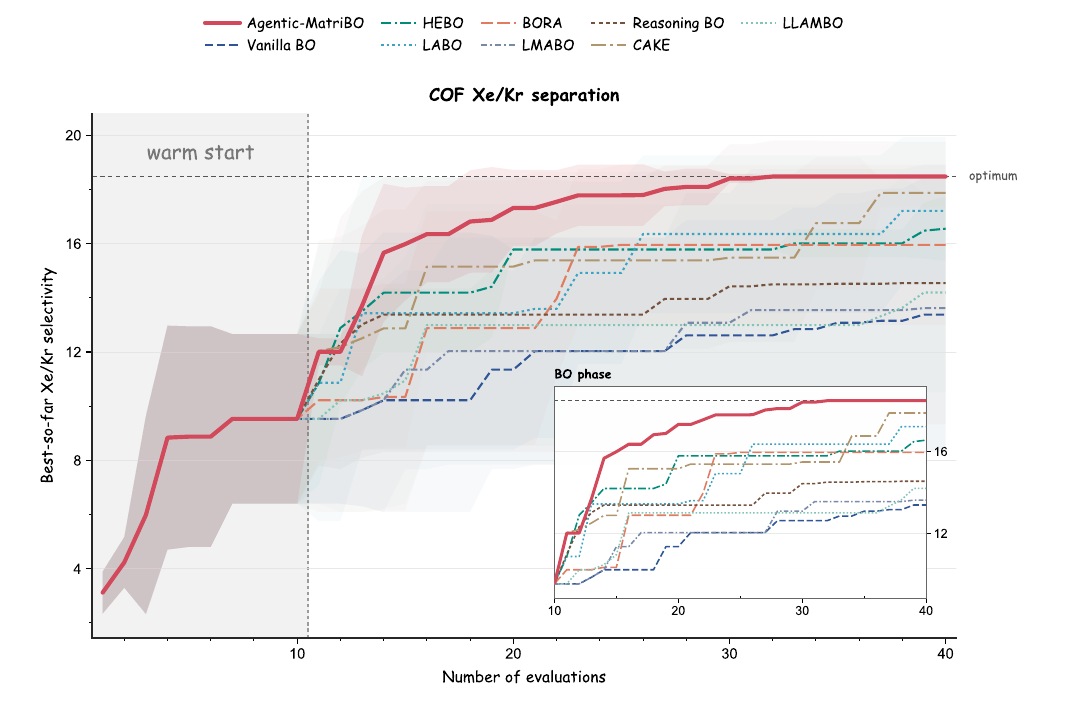}
\caption{\textbf{COF optimization with shared initialization.} For a given seed, the five methods in this group share the same 10 initial candidates and
observations, followed by 30 evaluations per method. Curves and shaded
regions show the five-seed mean and ±1 s.d.; the gray region denotes
initialization and the inset enlarges the subsequent stage.}\label{fig:shared-init}
\end{figure}

In the COF shared-initialization experiment, Agentic-MatriBO retains the
highest mean endpoint, BO-stage area, and late-stage area among the listed
methods, and all five seeds hit the tabulated optimum (Fig.~\ref{fig:shared-init}
and Table~\ref{tab:shared-init}). LMABO and BORA are included here because
LMABO is the strongest endpoint baseline in the end-to-end COF comparison;
the complete shared-initialization table is provided in Appendix A.2.

{
\footnotesize
\begin{longtable}[]{@{}
  >{\raggedright\arraybackslash}p{(\linewidth - 8\tabcolsep) * \real{0.2000}}
  >{\centering\arraybackslash}p{(\linewidth - 8\tabcolsep) * \real{0.2000}}
  >{\centering\arraybackslash}p{(\linewidth - 8\tabcolsep) * \real{0.2000}}
  >{\centering\arraybackslash}p{(\linewidth - 8\tabcolsep) * \real{0.2000}}
  >{\centering\arraybackslash}p{(\linewidth - 8\tabcolsep) * \real{0.2000}}@{}}
\caption[COF comparison with shared initialization]{\textbf{COF comparison with shared initialization.} Values are mean ± s.d. over five seeds. The five methods in this group receive the same ten initial candidates; the complete summary is in Appendix A.2. }\label{tab:shared-init}\\
\toprule\noalign{}
\begin{minipage}[c]{\linewidth}\raggedright
Method
\end{minipage} & \begin{minipage}[c]{\linewidth}\centering
Best-so-far ↑
\end{minipage} & \begin{minipage}[c]{\linewidth}\centering
nAUC-BO ↑
\end{minipage} & \begin{minipage}[c]{\linewidth}\centering
EF40 ↑
\end{minipage} & \begin{minipage}[c]{\linewidth}\centering
Late nAUC ↑
\end{minipage} \\
\midrule\noalign{}
\endfirsthead
\caption[]{\textbf{(continued)}}\\
\toprule\noalign{}
\begin{minipage}[c]{\linewidth}\raggedright
Method
\end{minipage} & \begin{minipage}[c]{\linewidth}\centering
Best-so-far ↑
\end{minipage} & \begin{minipage}[c]{\linewidth}\centering
nAUC-BO ↑
\end{minipage} & \begin{minipage}[c]{\linewidth}\centering
EF40 ↑
\end{minipage} & \begin{minipage}[c]{\linewidth}\centering
Late nAUC ↑
\end{minipage} \\
\midrule\noalign{}
\endhead
\bottomrule\noalign{}
\endlastfoot
HEBO & 16.550 ± 1.184 & 0.818 ± 0.032 & 0.896 ± 0.064 & 0.861 ± 0.010 \\
CAKE & 17.875 ± 0.531 & 0.823 ± 0.118 & 0.967 ± 0.029 & 0.868 ± 0.127 \\
LMABO & 13.618 ± 5.286 & 0.662 ± 0.224 & 0.737 ± 0.287 & 0.698 ± 0.256 \\
BORA & 15.953 ± 2.955 & 0.769 ± 0.107 & 0.863 ± 0.160 & 0.845 ± 0.148 \\
\mbox{\textbf{Agentic-MatriBO}} & \textbf{18.476 ± 0.000} & \textbf{0.923 ± 0.038} & \textbf{1.000 ± 0.000} & \textbf{0.982 ± 0.026} \\
\end{longtable}
}

Under identical initial conditions, Agentic-MatriBO has higher endpoint,
nAUC-BO, and late-stage area than CAKE. This is consistent with an effect
of subsequent search, although the experiment does not isolate individual
configurations.

\subsection{5.7 Grouped ablations separating starting point, recipe,
and
review}\label{grouped-ablations-separating-starting-point-recipe-and-review}

\begin{figure}[htbp]
\centering
\includegraphics[width=\linewidth,height=0.62\textheight,keepaspectratio]{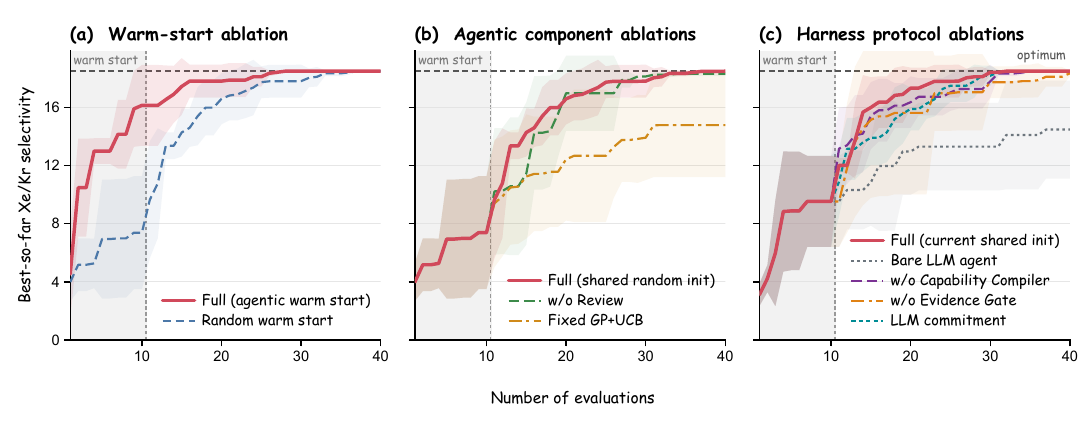}
\caption{\textbf{Grouped COF ablations.} (a) Warm-start
ablation: the complete system (agentic warm start) and random warm
start; (b) component ablation: the complete system, No review, and fixed
GP+UCB under shared random initialization; (c) Harness ablation: the
complete system, Bare LLM, No Capability Compiler, No Evidence
Gate, and LLM commitment under another shared initialization. Each
condition uses five seeds with a total budget of 10+30; lines and shaded
regions show the mean and ±1 s.d., and the dashed line denotes the
tabulated optimum. Each panel compares only against the complete system
in its own group.}\label{fig:ablations}
\end{figure}

The complete warm-start system and random-initialization variant reach
the same endpoint, but nAUC-BO is 0.970 versus 0.888 and \(t_{95}\) is
13.4 versus 23.4 (Fig.~\ref{fig:ablations}a and Table~\ref{tab:ablations}), showing the
benefit of the agentic starting point for early quality.

\clearpage
{
\footnotesize
\setlength{\tabcolsep}{3.5pt}
\begin{longtable}[]{@{}
  >{\raggedright\arraybackslash}p{(\linewidth - 12\tabcolsep) * \real{0.1300}}
  >{\raggedright\arraybackslash}p{(\linewidth - 12\tabcolsep) * \real{0.1500}}
  >{\centering\arraybackslash}p{(\linewidth - 12\tabcolsep) * \real{0.1700}}
  >{\centering\arraybackslash}p{(\linewidth - 12\tabcolsep) * \real{0.1500}}
  >{\centering\arraybackslash}p{(\linewidth - 12\tabcolsep) * \real{0.1500}}
  >{\centering\arraybackslash}p{(\linewidth - 12\tabcolsep) * \real{0.1200}}
  >{\centering\arraybackslash}p{(\linewidth - 12\tabcolsep) * \real{0.1300}}@{}}
\caption[Grouped COF ablation results]{\textbf{Grouped COF ablations.} Values are mean ± s.d. over five seeds; brackets in \(t_{95}\) give threshold-hit counts; --- indicates that exact-optimum hits were not separately recorded for that control. Comparisons are within groups, and ``Full'' denotes the complete system. Full names and results are in Appendix A.3.}\label{tab:ablations}\\
\toprule\noalign{}
\begin{minipage}[c]{\linewidth}\raggedright
Ablation group
\end{minipage} & \begin{minipage}[c]{\linewidth}\raggedright
Condition
\end{minipage} & \begin{minipage}[c]{\linewidth}\centering
Best-so-far ↑
\end{minipage} & \begin{minipage}[c]{\linewidth}\centering
nAUC-BO ↑
\end{minipage} & \begin{minipage}[c]{\linewidth}\centering
Late nAUC ↑
\end{minipage} & \begin{minipage}[c]{\linewidth}\centering
\(t_{95}\) ↓
\end{minipage} & \begin{minipage}[c]{\linewidth}\centering
Global optimum hits
\end{minipage} \\
\midrule\noalign{}
\endfirsthead
\caption[]{\textbf{(continued)}}\\
\toprule\noalign{}
\begin{minipage}[c]{\linewidth}\raggedright
Ablation group
\end{minipage} & \begin{minipage}[c]{\linewidth}\raggedright
Condition
\end{minipage} & \begin{minipage}[c]{\linewidth}\centering
Best-so-far ↑
\end{minipage} & \begin{minipage}[c]{\linewidth}\centering
nAUC-BO ↑
\end{minipage} & \begin{minipage}[c]{\linewidth}\centering
Late nAUC ↑
\end{minipage} & \begin{minipage}[c]{\linewidth}\centering
\(t_{95}\) ↓
\end{minipage} & \begin{minipage}[c]{\linewidth}\centering
Global optimum hits
\end{minipage} \\
\midrule\noalign{}
\endhead
\bottomrule\noalign{}
\endlastfoot
\multirow[t]{2}{*}{\textbf{Warm start}} & \textbf{Full (agentic)} & \textbf{\mbox{18.476 ± 0.000}} &
\textbf{\mbox{0.970 ± 0.039}} & \textbf{\mbox{0.991 ± 0.015}} & \textbf{\mbox{13.4 [5/5]}} & --- \\
& Random & \mbox{18.476 ± 0.000} & \mbox{0.888 ± 0.048} &
\mbox{0.967 ± 0.045} & \mbox{23.4 [5/5]} & --- \\
\addlinespace[3pt]
\multirow[t]{3}{*}{\textbf{Component}} & \textbf{Full} & \textbf{\mbox{18.476 ± 0.000}} & \textbf{\mbox{0.888 ± 0.048}} & \textbf{\mbox{0.967 ± 0.045}} & \textbf{\mbox{23.4 [5/5]}} & \textbf{5/5} \\
& No review & \mbox{18.300 ± 0.171} & \mbox{0.865 ± 0.082} & \mbox{0.963 ± 0.046} &
\mbox{20.0 [5/5]} & 0/5 \\
& Fixed GP+UCB & \mbox{14.773 ± 3.573} & \mbox{0.696 ± 0.189} & \mbox{0.753 ± 0.192} & \mbox{35.0 [2/5]} & --- \\
\addlinespace[3pt]
\multirow[t]{5}{*}{\textbf{Harness}} & \textbf{Full} & \textbf{\mbox{18.476 ± 0.000}} & \textbf{\mbox{0.923 ± 0.038}} & \textbf{\mbox{0.982 ± 0.026}} & \textbf{\mbox{19.6 [5/5]}} & \textbf{5/5} \\
& Bare LLM & \mbox{14.465 ± 3.365} & \mbox{0.694 ± 0.153} & \mbox{0.740 ± 0.170} & \mbox{34.2 [2/5]} & 0/5 \\
& w/o Compiler & \mbox{18.455 ± 0.028} & \mbox{0.903 ± 0.049} & \mbox{0.957 ± 0.032} & \mbox{23.6 [5/5]} & 3/5 \\
& w/o Evidence Gate & \mbox{18.330 ± 0.199} & \mbox{0.875 ± 0.160} & \mbox{0.936 ± 0.114} & \mbox{19.8 [5/5]} & 3/5 \\
& LLM Commitment & \mbox{18.476 ± 0.000} & \mbox{0.888 ± 0.045} & \mbox{0.963 ± 0.028} & \mbox{25.0 [5/5]} & 5/5 \\
\end{longtable}
}

Within the component group, fixed GP+UCB reduces the endpoint and
nAUC-BO relative to the complete system, and only 2/5 seeds reach the
95\% threshold. This control characterizes the overall difference after
removing online recipe adaptation.

The No review variant has a lower endpoint and cumulative area, yet its
\(t_{95}\) decreases from 23.4 to 20.0 and its global-optimum hit count
decreases from 5/5 to 0/5. Thus, crossing the 95\% threshold earlier
does not necessarily imply higher endpoint quality or a higher exact-hit
rate.

\subsection{5.8 Effects of capability compilation, evidence gating, and
candidate commitment}\label{capability-compilation-evidence-gating-and-commitment-configuration}

In the Harness group, removing the capability compiler or evidence gate
reduces cumulative area and exact-optimum hits, although all three
conditions reach the 95\% threshold in 5/5 seeds. The complete system
has the largest cumulative area.

LLM commitment preserves the endpoint and exact-hit count but lowers
nAUC-BO and delays \(t_{95}\); Bare LLM performs substantially worse and
does not hit the global optimum in any seed.

\subsection{5.9 Observed model differences in initialization and early area}\label{model-differences-concentrate-in-initialization-and-early-area}

The model participates in initialization and subsequent decisions.
In these five-seed runs, the largest observed between-model differences occur during initialization and evaluations 10--20 (Fig.~\ref{fig:model-sensitivity} and Table~\ref{tab:models}); model-call mappings are
provided in Appendix A.3.

\begin{figure}[htbp]
\centering
\includegraphics[width=\linewidth,height=0.62\textheight,keepaspectratio]{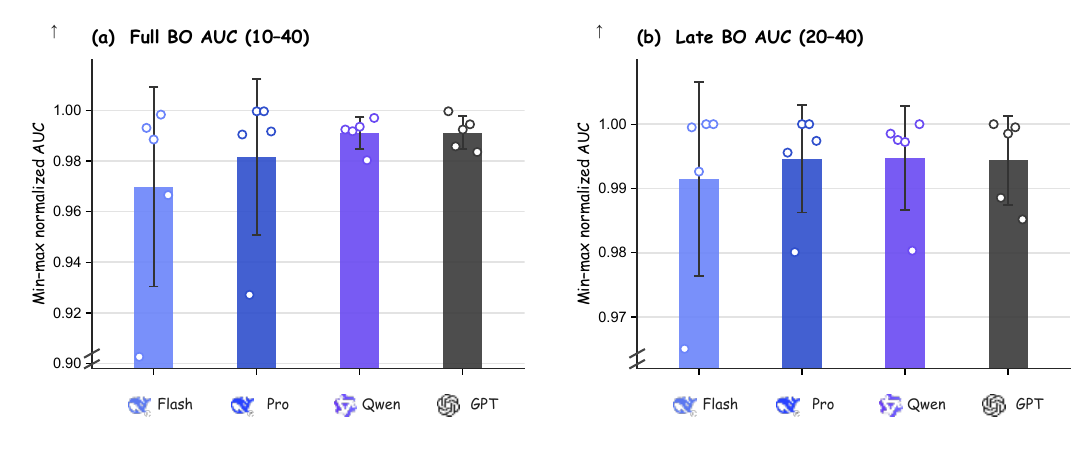}
\caption{\textbf{LLM sensitivity on COF.} (a) Area over
evaluations 10--40; (b) late-stage area over evaluations 20--40. Bars,
points, and error bars show the five-seed mean, individual-seed values,
and ±1 s.d., respectively. The y-axis uses a truncated scale. This
comparison includes the effect of the model on both initialization and
subsequent strategy decisions.}\label{fig:model-sensitivity}
\end{figure}

{
\footnotesize
\begin{longtable}[]{@{}
  >{\raggedright\arraybackslash}p{(\linewidth - 10\tabcolsep) * \real{0.2000}}
  >{\centering\arraybackslash}p{(\linewidth - 10\tabcolsep) * \real{0.1500}}
  >{\centering\arraybackslash}p{(\linewidth - 10\tabcolsep) * \real{0.1600}}
  >{\centering\arraybackslash}p{(\linewidth - 10\tabcolsep) * \real{0.1700}}
  >{\centering\arraybackslash}p{(\linewidth - 10\tabcolsep) * \real{0.1700}}
  >{\centering\arraybackslash}p{(\linewidth - 10\tabcolsep) * \real{0.1500}}@{}}
\caption[COF model comparison]{\textbf{COF model comparison.} Initialization best
value, BO-stage area, and late-stage nAUC are reported as mean ± s.d.;
early nAUC is reported as a mean; \(n=5\) per condition. The complete
summary is provided in Appendix A.3.}\label{tab:models}\\
\toprule\noalign{}
\begin{minipage}[c]{\linewidth}\raggedright
Model
\end{minipage} & \begin{minipage}[c]{\linewidth}\centering
Init. best
\end{minipage} & \begin{minipage}[c]{\linewidth}\centering
Early nAUC (10--20)
\end{minipage} & \begin{minipage}[c]{\linewidth}\centering
nAUC-BO (10--40)
\end{minipage} & \begin{minipage}[c]{\linewidth}\centering
Late nAUC (20--40)
\end{minipage} & \begin{minipage}[c]{\linewidth}\centering
Optimum hits
\end{minipage} \\
\midrule\noalign{}
\endfirsthead
\caption[]{\textbf{(continued)}}\\
\toprule\noalign{}
\begin{minipage}[c]{\linewidth}\raggedright
Model
\end{minipage} & \begin{minipage}[c]{\linewidth}\centering
Init. best
\end{minipage} & \begin{minipage}[c]{\linewidth}\centering
Early nAUC (10--20)
\end{minipage} & \begin{minipage}[c]{\linewidth}\centering
nAUC-BO (10--40)
\end{minipage} & \begin{minipage}[c]{\linewidth}\centering
Late nAUC (20--40)
\end{minipage} & \begin{minipage}[c]{\linewidth}\centering
Optimum hits
\end{minipage} \\
\midrule\noalign{}
\endhead
\bottomrule\noalign{}
\endlastfoot
DeepSeek 4.0 Flash & \mbox{16.128 ± 2.777} & 0.9266 & \mbox{0.9698 ± 0.0395} & \mbox{0.9914 ± 0.0151} & 5/5 \\
DeepSeek 4.0 Pro & \mbox{17.281 ± 1.860} & 0.9560 & \mbox{0.9817 ± 0.0308} & \mbox{0.9946 ± 0.0083} & 5/5 \\
Qwen3-Max & \mbox{18.113 ± 0.000} & 0.9837 & \mbox{0.9910 ± 0.0063} & \mbox{0.9947 ± 0.0081} & 3/5 \\
GPT-5.6-sol & \mbox{18.113 ± 0.000} & 0.9849 & \mbox{0.9912 ± 0.0066} & \mbox{0.9944 ± 0.0070} & 5/5 \\
\end{longtable}
}

Late-stage area means for the strongest models differ by only about
0.0003, while BO-stage differences occur mainly in evaluations 10--20
(Appendix B.2.4). Cumulative near-optimal quality and exact
hits therefore capture different search characteristics.

Because these model-dependent differences could also reflect prior exposure
to public candidate tables, Section 5.10 separately audits whether original
candidate IDs provide a consistent ranking advantage. This audit addresses
a possible confound rather than serving as another optimization comparison.

\subsection{5.10 Zero-observation ranking shows no consistent hit
advantage}\label{zero-observation-ranking-shows-no-consistent-hit-advantage}

BH and Suzuki use public reaction-yield tables
\cite{ahneman2018reaction,perera2018platform}. The audit evaluates rankings made without target values and tests whether
original IDs affect hit rates. Each trial
contains one tabulated optimum and 149 randomly sampled non-optimal
candidates; models return a top-40 list from the same descriptors and
ordering under original or re-encoded IDs. Tied Suzuki optima are
rotated as anchors. Each ID group is repeated 10 times for four tasks and
four models in paired trials at temperature 0.7. The corresponding random
top-40 hit probability is 40/150 (26.7\%).

COF IDs are public structure filenames, whereas the other IDs were
assigned during table construction; the ID-control interpretation is
therefore strongest for COF. All ranking requests were independent of
the Agentic-MatriBO search process.

\begin{figure}[H]
\centering
\includegraphics[width=\linewidth,height=0.52\textheight,keepaspectratio]{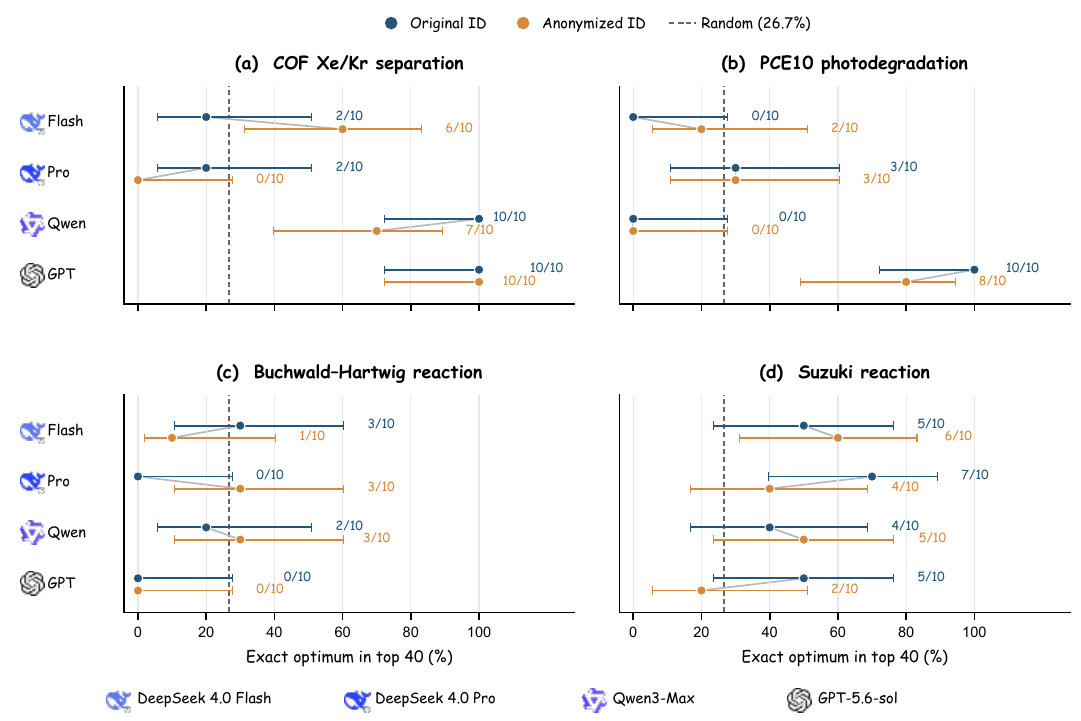}
\caption{\textbf{Original and anonymous IDs in zero-observation
ranking.} (a) COF; (b) PCE10; (c) BH; (d) Suzuki. Each row corresponds
to a model. Blue points denote original IDs and orange points anonymous
IDs; the horizontal axis is the fraction of the anchored ID entering the
top 40 across 10 rankings, error bars are Wilson 95\% intervals, and
numbers are hit counts. The vertical dashed line is the random level of
26.7\%. Models receive candidate IDs and descriptors directly, without
the Agentic-MatriBO strategy or execution process.}\label{fig:id-audit}
\end{figure}

Across the 16 model--task pairs, original-ID hit rates were higher,
lower, or equal in 6, 6, and 4 pairs, with aggregate hits of 63/160 and
60/160. After Benjamini--Hochberg correction, paired McNemar and sign
tests both give \(q=1.0\); original IDs therefore show no consistent hit
advantage.

The audit does not establish equivalence: ten paired replicates limit
power, and both groups retain descriptors, so descriptor knowledge is not
isolated.

\section{6. Discussion}\label{discussion}

\subsection{6.1 Capability compilation connects task evidence and
algorithm
composition}\label{capability-compilation-connects-task-evidence-and-algorithm-composition}

Under the target protocol, capability compilation exposes component
applicability conditions to the strategy and uses them for
proposal-admission checks. The language
strategy uses task semantics, diagnostics, and history to propose
recipes and candidate plans, while the execution layer performs
numerical computation and candidate submission; domain information
enters the next selection through configuration and ranking. In the 20
runs shown in Figure~\ref{fig:C1}, the effective surrogate model
switched at least once in every run, whereas acquisition-function and
search-strategy switches were less frequent, with 9 runs retaining each
of these components unchanged throughout.

\subsection{6.2 Gains depend on task, stage, and
configuration}\label{gains-depend-on-task-stage-and-configuration}

Endpoint quality and cumulative area answer different search questions:
the former measures the best candidate found by the end of the budget,
whereas the latter also reflects when high-quality candidates first
appear. The PCE10 and Suzuki results show that endpoint improvement can
accompany a very small area gain, or even a slightly lower cumulative
area. The COF No review variant crossed the 95\% threshold earlier but
did not hit the exact optimum, further showing that threshold attainment
and exact hits should be evaluated separately.

Shared initialization focuses the comparison on subsequent search,
whereas the warm-start control shows that initial candidates can affect
cumulative quality. Fixed recipes, No review, and LLM commitment alter
the search process at different points, and the results vary with both
starting points and subsequent configurations; their individual
contributions cannot be separated from these comparisons. The
contributions of candidate distribution, component adaptation, and
domain knowledge have not been independently separated.

\subsection{6.3 Evidence levels in cross-round
feedback}\label{evidence-levels-in-cross-round-feedback}

Research memory links prior expectations to observations and stores
diagnostics, reflections, and subsequent evidence requirements for
retrieval by the next-round strategy. It provides cross-round context
for compositional adaptation. The current experiments do not include an
independent no-memory or no-Playbook control, so the standalone gain of
the memory module cannot yet be quantified.

\section{7. Limitations}\label{limitations}

\textbf{Evaluation scope and statistical analysis.} The evaluation
covers four offline finite candidate pools with five seeds per condition;
COF shared-initialization and mechanism ablations were not repeated on
the other tasks. The tabulated optimum is defined relative to each
candidate set, so results may not generalize to open-ended generation,
continuous multi-objective settings, or asynchronous noisy experiments.
Offline search and path projections cannot validate materials mechanisms
or replace prospective wet-lab experiments. We report means, standard
deviations, and exact-hit counts descriptively; five seeds limit the
precision of between-condition estimates.

\textbf{Public data and model priors.} In the zero-observation audit,
both original-ID and re-encoded-ID prompts retain reagent names and
structural descriptors. The BH and Suzuki identifiers were assigned
during table construction, but the audit cannot separate general chemical
knowledge from memorization of reaction conditions in public yield tables.
This limits interpretation of those two tasks.

\textbf{Implementation coverage.} The evaluated code mainly checks
newly switched components on the CHANGE path. It does not yet verify that
cited support materials substantiate their premises, revalidate deferred
components after CHANGE-to-KEEP transitions, or apply the full check set
on heuristic paths. The empty-candidate path also raises a selection
error rather than following the declared no-candidate termination path.
The formal properties in Section 4.7 are conditional protocol properties,
not guarantees that the current implementation enforces every check; see
Section 4.8 and Appendix C.4.

\clearpage

\section{8. Conclusion}\label{conclusion}

Across four offline finite candidate pools, eight baselines, and five
seeds per condition, Agentic-MatriBO has the best mean endpoint quality
in each task's objective direction, while cumulative-quality gains
depend on the task. After shared initialization on COF, all five seeds
reached the tabulated optimum while retaining an area advantage,
indicating that differences in this group also arise from subsequent
search. Ablations show that threshold attainment, cumulative quality, and
exact hits respond differently to configuration. Observed between-model
area differences are largest during initialization and the early stage.
The zero-observation audit found no consistent hit advantage from
original candidate IDs.

Agentic-MatriBO registers component-selection conditions as capability
contracts and uses evidence-aware compilation and compositional adaptation
to connect language strategy with numerical search. Under the declared
protocol, the execution layer validates and commits candidates, and oracle
observations update research memory for later rounds. The evaluated
implementation covers the protocol only partially, as detailed in Section
4.8 and Appendix C.4.

\bibliographystyle{unsrt}
{\small
\bibliography{references}
}

\section{Declarations}\label{declarations}

\textbf{Data and Code Availability.}
The study uses four finite candidate pools: COF, PCE10,
Buchwald--Hartwig, and Suzuki--Miyaura. Their sources and processing are
described in Appendices B.2.3 and B.2.6. The benchmark tables and
source-processing details can be provided by the authors upon reasonable
request. The complete implementation code and per-seed result files are not
publicly released at this stage.

\textbf{Ethics Declaration.}
This software and finite-table benchmark study involve no human
participants, human data, animals, or animal-derived samples;
institutional approval for human or animal subjects is therefore not
applicable.

\textbf{Author Contributions.}
All authors contributed to the study design, methodology, analysis, and
manuscript preparation. All authors reviewed and approved the final
manuscript.

\textbf{Competing Interests.}
The authors declare no competing interests.

\textbf{Funding.}
No external funding was declared for this work.

\clearpage
\appendix
\setcounter{secnumdepth}{3}
\setcounter{figure}{0}
\setcounter{table}{0}
\makeatletter
\@addtoreset{figure}{section}
\@addtoreset{table}{section}
\renewcommand{\thefigure}{\Alph{section}\arabic{figure}}
\renewcommand{\thetable}{\Alph{section}\arabic{table}}
\renewcommand{\theHfigure}{\Alph{section}\arabic{figure}}
\renewcommand{\theHtable}{\Alph{section}\arabic{table}}
\makeatother
\begingroup
\AtBeginEnvironment{longtable}{\footnotesize\setlength{\tabcolsep}{1.5pt}}
\renewcommand{\labelitemi}{\ensuremath{\bullet}}
This supplement reports the values underlying the main-text figures and
tables, along with process cases, metric definitions, and implementation
details.

\section{Detailed Experimental Results}\label{app:A}

\subsection{Values underlying Fig.~3 and Table~3}\label{app:A.1}

This table reports the five-seed means underlying Fig.~3 and Table~3 for
a shared oracle budget of 10 initialization evaluations and 30 subsequent
evaluations. Arrows on task names indicate the direction of the original
objective. nAUC-BO, EF40, and late-stage nAUC are larger-is-better.
Improv.\ denotes the percentage change relative to the best baseline for
the same task and metric; positive values favor the proposed method.

{
\begin{longtable}{@{}>{\raggedright\arraybackslash}p{0.11\linewidth}>{\raggedright\arraybackslash}p{0.18\linewidth}>{\centering\arraybackslash}p{0.167\linewidth}>{\centering\arraybackslash}p{0.167\linewidth}>{\centering\arraybackslash}p{0.167\linewidth}>{\centering\arraybackslash}p{0.167\linewidth}@{}}
\caption[Complete baseline comparisons across the four tasks]{\textbf{Complete baseline comparisons across the four tasks.} Values are mean $\pm$ s.d. over five seeds per condition. Rows list one method per line; the final row of each task block reports the improvement of Agentic-MatriBO relative to the best baseline.}\label{tab:s-baselines}\\
\toprule
Dataset & Method & Best-so-far & nAUC-BO & EF40 & Late-stage nAUC \\
\midrule
\endfirsthead
\caption[]{\textbf{(continued)}}\\
\toprule
Dataset & Method & Best-so-far & nAUC-BO & EF40 & Late-stage nAUC \\
\midrule
\endhead
\bottomrule
\endlastfoot
COF $\uparrow$ & Vanilla BO & $13.381\,\pm\,5.055$ & $0.636\,\pm\,0.206$ & $0.724\,\pm\,0.274$ & $0.679\,\pm\,0.236$ \\
 & HEBO & $16.550\,\pm\,1.184$ & $0.818\,\pm\,0.032$ & $0.896\,\pm\,0.064$ & $0.861\,\pm\,0.010$ \\
 & LABO & $15.564\,\pm\,3.148$ & $0.683\,\pm\,0.203$ & $0.842\,\pm\,0.171$ & $0.739\,\pm\,0.168$ \\
 & CAKE & $17.004\,\pm\,3.289$ & $0.847\,\pm\,0.164$ & $0.920\,\pm\,0.178$ & $0.877\,\pm\,0.166$ \\
 & LMABO & $17.326\,\pm\,1.416$ & $0.868\,\pm\,0.076$ & $0.938\,\pm\,0.077$ & $0.898\,\pm\,0.075$ \\
 & BORA & $16.723\,\pm\,1.919$ & $0.832\,\pm\,0.135$ & $0.905\,\pm\,0.104$ & $0.863\,\pm\,0.137$ \\
 & Reasoning BO & $17.089\,\pm\,1.349$ & $0.801\,\pm\,0.154$ & $0.925\,\pm\,0.073$ & $0.871\,\pm\,0.099$ \\
 & LLAMBO & $16.329\,\pm\,1.891$ & $0.794\,\pm\,0.134$ & $0.884\,\pm\,0.103$ & $0.821\,\pm\,0.116$ \\
 & Agentic-MatriBO & $18.476\,\pm\,0.000$ & $0.970\,\pm\,0.039$ & $1.000\,\pm\,0.000$ & $0.991\,\pm\,0.015$ \\
 & \textit{Improvement} & $6.63\%$ & $11.71\%$ & $6.65\%$ & $10.37\%$ \\
\addlinespace[2pt]
PCE10 $\downarrow$ & Vanilla BO & $0.0070\,\pm\,0.0049$ & $0.980\,\pm\,0.016$ & $0.993\,\pm\,0.007$ & $0.991\,\pm\,0.009$ \\
 & HEBO & $0.0217\,\pm\,0.0258$ & $0.963\,\pm\,0.033$ & $0.973\,\pm\,0.035$ & $0.968\,\pm\,0.034$ \\
 & LABO & $0.0086\,\pm\,0.0066$ & $0.986\,\pm\,0.013$ & $0.991\,\pm\,0.009$ & $0.991\,\pm\,0.009$ \\
 & CAKE & $0.0034\,\pm\,0.0040$ & $0.974\,\pm\,0.024$ & $0.998\,\pm\,0.005$ & $0.991\,\pm\,0.010$ \\
 & LMABO & $0.0235\,\pm\,0.0053$ & $0.967\,\pm\,0.008$ & $0.970\,\pm\,0.007$ & $0.970\,\pm\,0.007$ \\
 & BORA & $0.0161\,\pm\,0.0096$ & $0.967\,\pm\,0.019$ & $0.981\,\pm\,0.013$ & $0.975\,\pm\,0.014$ \\
 & Reasoning BO & $0.0040\,\pm\,0.0039$ & $0.9963\,\pm\,0.0062$ & $0.997\,\pm\,0.005$ & $0.997\,\pm\,0.005$ \\
 & LLAMBO & $0.0040\,\pm\,0.0039$ & $0.986\,\pm\,0.005$ & $0.997\,\pm\,0.005$ & $0.994\,\pm\,0.005$ \\
 & Agentic-MatriBO & $0.00162\,\pm\,0.00000$ & $0.9964\,\pm\,0.0021$ & $1.000\,\pm\,0.000$ & $1.000\,\pm\,0.000$ \\
 & \textit{Improvement} & $52.35\%$ & $0.01\%$ & $0.24\%$ & $0.31\%$ \\
\addlinespace[2pt]
BH $\uparrow$ & Vanilla BO & $85.644\,\pm\,1.000$ & $0.951\,\pm\,0.048$ & $0.989\,\pm\,0.012$ & $0.961\,\pm\,0.028$ \\
 & HEBO & $83.873\,\pm\,1.470$ & $0.960\,\pm\,0.019$ & $0.969\,\pm\,0.017$ & $0.964\,\pm\,0.018$ \\
 & LABO & $83.556\,\pm\,1.865$ & $0.950\,\pm\,0.013$ & $0.965\,\pm\,0.022$ & $0.959\,\pm\,0.015$ \\
 & CAKE & $82.004\,\pm\,1.156$ & $0.915\,\pm\,0.038$ & $0.947\,\pm\,0.013$ & $0.928\,\pm\,0.028$ \\
 & LMABO & $81.541\,\pm\,3.349$ & $0.928\,\pm\,0.042$ & $0.942\,\pm\,0.039$ & $0.938\,\pm\,0.046$ \\
 & BORA & $84.400\,\pm\,1.706$ & $0.963\,\pm\,0.008$ & $0.975\,\pm\,0.020$ & $0.971\,\pm\,0.016$ \\
 & Reasoning BO & $84.926\,\pm\,1.355$ & $0.956\,\pm\,0.024$ & $0.981\,\pm\,0.016$ & $0.973\,\pm\,0.015$ \\
 & LLAMBO & $82.950\,\pm\,3.354$ & $0.922\,\pm\,0.063$ & $0.958\,\pm\,0.039$ & $0.938\,\pm\,0.060$ \\
 & Agentic-MatriBO & $86.316\,\pm\,0.385$ & $0.963\,\pm\,0.011$ & $0.997\,\pm\,0.004$ & $0.981\,\pm\,0.015$ \\
 & \textit{Improvement} & $0.78\%$ & $-0.04\%$ & $0.79\%$ & $0.89\%$ \\
\addlinespace[2pt]
Suzuki $\uparrow$ & Vanilla BO & $95.352\,\pm\,1.271$ & $0.915\,\pm\,0.049$ & $0.954\,\pm\,0.013$ & $0.931\,\pm\,0.042$ \\
 & HEBO & $94.982\,\pm\,3.728$ & $0.919\,\pm\,0.030$ & $0.951\,\pm\,0.037$ & $0.943\,\pm\,0.035$ \\
 & LABO & $95.998\,\pm\,0.967$ & $0.920\,\pm\,0.048$ & $0.961\,\pm\,0.010$ & $0.950\,\pm\,0.018$ \\
 & CAKE & $96.990\,\pm\,1.580$ & $0.911\,\pm\,0.107$ & $0.971\,\pm\,0.016$ & $0.937\,\pm\,0.076$ \\
 & LMABO & $92.500\,\pm\,5.889$ & $0.909\,\pm\,0.051$ & $0.926\,\pm\,0.059$ & $0.918\,\pm\,0.054$ \\
 & BORA & $96.950\,\pm\,2.303$ & $0.950\,\pm\,0.028$ & $0.970\,\pm\,0.023$ & $0.958\,\pm\,0.030$ \\
 & Reasoning BO & $94.342\,\pm\,7.906$ & $0.935\,\pm\,0.074$ & $0.944\,\pm\,0.079$ & $0.941\,\pm\,0.078$ \\
 & LLAMBO & $93.618\,\pm\,1.532$ & $0.881\,\pm\,0.072$ & $0.937\,\pm\,0.015$ & $0.899\,\pm\,0.049$ \\
 & Agentic-MatriBO & $97.558\,\pm\,1.038$ & $0.946\,\pm\,0.033$ & $0.977\,\pm\,0.010$ & $0.965\,\pm\,0.020$ \\
 & \textit{Improvement} & $0.59\%$ & $-0.51\%$ & $0.59\%$ & $0.74\%$ \\
\end{longtable}
}

\subsection{Values underlying Fig.~4 and Table~4}\label{app:A.2}

This table records the five-seed means behind Fig.~4 and Table~4. All
methods in a seed share the same 10 initial observations; the COF rows
in Appendix A.1 instead use each method's own initialization. The reading rules
for arrows and Improv.\ are those in Appendix A.1.

{
\begin{longtable}{@{}>{\raggedright\arraybackslash}p{0.11\linewidth}>{\raggedright\arraybackslash}p{0.18\linewidth}>{\centering\arraybackslash}p{0.167\linewidth}>{\centering\arraybackslash}p{0.167\linewidth}>{\centering\arraybackslash}p{0.167\linewidth}>{\centering\arraybackslash}p{0.167\linewidth}@{}}
\caption[Complete COF comparison with shared initialization]{\textbf{Complete COF comparison with shared initialization.} Values are mean $\pm$ s.d. over five seeds per condition. Rows list one method per line; the final row of each task block reports the improvement of Agentic-MatriBO relative to the best baseline.}\label{tab:s-shared-init}\\
\toprule
Dataset & Method & Best-so-far & nAUC-BO & EF40 & Late-stage nAUC \\
\midrule
\endfirsthead
\caption[]{\textbf{(continued)}}\\
\toprule
Dataset & Method & Best-so-far & nAUC-BO & EF40 & Late-stage nAUC \\
\midrule
\endhead
\bottomrule
\endlastfoot
COF $\uparrow$ & Vanilla BO & $13.381\,\pm\,5.055$ & $0.636\,\pm\,0.206$ & $0.724\,\pm\,0.274$ & $0.679\,\pm\,0.236$ \\
 & HEBO & $16.550\,\pm\,1.184$ & $0.818\,\pm\,0.032$ & $0.896\,\pm\,0.064$ & $0.861\,\pm\,0.010$ \\
 & LABO & $17.209\,\pm\,2.718$ & $0.803\,\pm\,0.163$ & $0.931\,\pm\,0.147$ & $0.860\,\pm\,0.148$ \\
 & CAKE & $17.875\,\pm\,0.531$ & $0.823\,\pm\,0.118$ & $0.967\,\pm\,0.029$ & $0.868\,\pm\,0.127$ \\
 & LMABO & $13.618\,\pm\,5.286$ & $0.662\,\pm\,0.224$ & $0.737\,\pm\,0.287$ & $0.698\,\pm\,0.256$ \\
 & BORA & $15.953\,\pm\,2.955$ & $0.769\,\pm\,0.107$ & $0.863\,\pm\,0.160$ & $0.845\,\pm\,0.148$ \\
 & Reasoning BO & $14.541\,\pm\,5.212$ & $0.738\,\pm\,0.266$ & $0.787\,\pm\,0.283$ & $0.760\,\pm\,0.276$ \\
 & LLAMBO & $14.197\,\pm\,3.676$ & $0.680\,\pm\,0.217$ & $0.768\,\pm\,0.199$ & $0.710\,\pm\,0.239$ \\
 & Agentic-MatriBO & $18.476\,\pm\,0.000$ & $0.923\,\pm\,0.038$ & $1.000\,\pm\,0.000$ & $0.982\,\pm\,0.026$ \\
 & \textit{Improvement} & $3.36\%$ & $12.24\%$ & $3.36\%$ & $13.11\%$ \\
\end{longtable}
}

\subsection{Ablation and Model Comparisons}\label{app:A.3}

Each ablation group is compared with the complete system in that group.
\(t_{95}\) is averaged over all five seeds according to Appendix B.2.2; the number
in brackets is the number of seeds that reach \(Q_n\ge0.95\) within
the budget. The random-initialization warm-start group and the full
system in the component group have identical numerical values for
Best-so-far, nAUC-BO, EF40, late-stage nAUC, and \(t_{95}\), and are not
treated as two independent repetitions; the global optimum hit in Table
5 of the main text is reported only for the component group. The COF row
in Table 3 is identical to the complete warm-start system; Table 4 and
the complete Harness group both report 0.923 under shared
initialization.

{
\begin{longtable}[]{@{}
  >{\raggedright\arraybackslash}p{(\linewidth - 12\tabcolsep) * \real{0.1500}}
  >{\raggedright\arraybackslash}p{(\linewidth - 12\tabcolsep) * \real{0.1700}}
  >{\centering\arraybackslash}p{(\linewidth - 12\tabcolsep) * \real{0.1400}}
  >{\centering\arraybackslash}p{(\linewidth - 12\tabcolsep) * \real{0.1200}}
  >{\centering\arraybackslash}p{(\linewidth - 12\tabcolsep) * \real{0.1200}}
  >{\centering\arraybackslash}p{(\linewidth - 12\tabcolsep) * \real{0.1300}}
  >{\centering\arraybackslash}p{(\linewidth - 12\tabcolsep) * \real{0.1700}}@{}}
\caption[Complete grouped COF ablation results]{\textbf{Complete grouped COF ablation results.} Each condition uses five seeds; brackets after $t_{95}$ report n/5, the number of seeds reaching the threshold out of five.}\label{tab:s-ablations}\\
\toprule
\noalign{}
\begin{minipage}[c]{\linewidth}\raggedright
Ablation group
\end{minipage} & \begin{minipage}[c]{\linewidth}\raggedright
Condition
\end{minipage} & \begin{minipage}[c]{\linewidth}\centering
Best-so-far ↑
\end{minipage} & \begin{minipage}[c]{\linewidth}\centering
nAUC-BO ↑
\end{minipage} & \begin{minipage}[c]{\linewidth}\centering
EF40 ↑
\end{minipage} & \begin{minipage}[c]{\linewidth}\centering
Late nAUC ↑
\end{minipage} & \begin{minipage}[c]{\linewidth}\centering
\(t_{95}\) ↓
\end{minipage} \\
\midrule
\noalign{}
\endfirsthead
\caption[]{\textbf{(continued)}}\\
\toprule
\noalign{}
\begin{minipage}[c]{\linewidth}\raggedright
Ablation group
\end{minipage} & \begin{minipage}[c]{\linewidth}\raggedright
Condition
\end{minipage} & \begin{minipage}[c]{\linewidth}\centering
Best-so-far ↑
\end{minipage} & \begin{minipage}[c]{\linewidth}\centering
nAUC-BO ↑
\end{minipage} & \begin{minipage}[c]{\linewidth}\centering
EF40 ↑
\end{minipage} & \begin{minipage}[c]{\linewidth}\centering
Late nAUC ↑
\end{minipage} & \begin{minipage}[c]{\linewidth}\centering
\(t_{95}\) ↓
\end{minipage} \\
\midrule
\noalign{}
\endhead
\bottomrule
\noalign{}
\endlastfoot
\multirow[t]{2}{*}{\textbf{Warm start}} & \textbf{Full (agentic)} & \textbf{\(18.476\,\pm\,0.000\)} & \textbf{\(0.970\,\pm\,0.039\)} &
\textbf{\(1.000\,\pm\,0.000\)} & \textbf{\(0.991\,\pm\,0.015\)} & \textbf{13.4 [5/5]} \\
& Random & \(18.476\,\pm\,0.000\) & \(0.888\,\pm\,0.048\) & \(1.000\,\pm\,0.000\) &
\(0.967\,\pm\,0.045\) & 23.4 [5/5] \\
\addlinespace[3pt]
\multirow[t]{3}{*}{\textbf{Component}} & \textbf{Full} & \textbf{\(18.476\,\pm\,0.000\)} &
\textbf{\(0.888\,\pm\,0.048\)} & \textbf{\(1.000\,\pm\,0.000\)} & \textbf{\(0.967\,\pm\,0.045\)} & \textbf{23.4 [5/5]} \\
& No review & \(18.300\,\pm\,0.171\) & \(0.865\,\pm\,0.082\) & \(0.990\,\pm\,0.009\) & \(0.963\,\pm\,0.046\) &
20.0 [5/5] \\
& Fixed GP+UCB & \(14.773\,\pm\,3.573\) & \(0.696\,\pm\,0.189\) & \(0.799\,\pm\,0.194\) & \(0.753\,\pm\,0.192\)
& 35.0 [2/5] \\
\addlinespace[3pt]
\multirow[t]{5}{*}{\textbf{Harness}} & \textbf{Full} & \textbf{\(18.476\,\pm\,0.000\)} &
\textbf{\(0.923\,\pm\,0.038\)} & \textbf{\(1.000\,\pm\,0.000\)} & \textbf{\(0.982\,\pm\,0.026\)} & \textbf{19.6 [5/5]} \\
& Bare LLM & \(14.465\,\pm\,3.365\) & \(0.694\,\pm\,0.153\) & \(0.783\,\pm\,0.182\) &
\(0.740\,\pm\,0.170\) & 34.2 [2/5] \\
& w/o Compiler & \(18.455\,\pm\,0.028\) & \(0.903\,\pm\,0.049\) & \(0.999\,\pm\,0.002\) &
\(0.957\,\pm\,0.032\) & 23.6 [5/5] \\
& w/o Evidence Gate & \(18.330\,\pm\,0.199\) & \(0.875\,\pm\,0.160\) & \(0.992\,\pm\,0.011\) &
\(0.936\,\pm\,0.114\) & 19.8 [5/5] \\
& LLM Commitment & \(18.476\,\pm\,0.000\) & \(0.888\,\pm\,0.045\) & \(1.000\,\pm\,0.000\) &
\(0.963\,\pm\,0.028\) & 25.0 [5/5] \\
\end{longtable}
}

The initialization best value, BO-stage area, and late-stage area are
reported as mean ± s.d.; early-stage area is reported as a mean. The
main text and tables use DeepSeek 4.0 Flash, DeepSeek 4.0 Pro,
Qwen3-Max, and GPT-5.6-sol, whose invocation names are DeepSeek-V4-Flash,
DeepSeek-V4-Pro, qwen3-max, and GPT-5.6-sol, respectively. The legend
names DeepSeek 4.0 Flash and DeepSeek 4.0 Pro are the first two models;
the remaining names are identical to those in the main text.

The model-sensitivity records use the invocation labels
DeepSeek-V4-Flash, DeepSeek-V4-Pro, qwen3-max, and GPT-5.6-sol; the
implementation specifies Vendor3/DeepSeek-V4-Flash as the default for
enabled LLM roles. These labels are invocation names rather than
immutable public snapshot IDs. Policy and reviewer calls use temperature
0.0, while initializer, surrogate, candidate-sampler, and low-fidelity
calls use temperature 0.7. The zero-observation ranking audit uses
temperature 0.7. The corresponding maximum output lengths are 8,192 tokens
for policy, reviewer, and candidate-sampler calls, 4,096 for surrogate
calls, and 2,048 for initializer and low-fidelity calls. Exact call counts,
access dates, and per-call token or cost records were not retained in the
aggregate result tables; therefore LLM cost and call overhead are not
compared quantitatively.

{
\begin{longtable}[]{@{}
  >{\raggedright\arraybackslash}p{(\linewidth - 10\tabcolsep) * \real{0.1667}}
  >{\centering\arraybackslash}p{(\linewidth - 10\tabcolsep) * \real{0.1667}}
  >{\centering\arraybackslash}p{(\linewidth - 10\tabcolsep) * \real{0.1667}}
  >{\centering\arraybackslash}p{(\linewidth - 10\tabcolsep) * \real{0.1667}}
  >{\centering\arraybackslash}p{(\linewidth - 10\tabcolsep) * \real{0.1667}}
  >{\raggedright\arraybackslash}p{(\linewidth - 10\tabcolsep) * \real{0.1667}}@{}}
\caption[Complete COF model comparison]{\textbf{Complete COF model comparison.} Each condition uses five seeds; the final column reports exact-optimum hits and evaluation indices.}\label{tab:s-models}\\
\toprule
\noalign{}
\begin{minipage}[c]{\linewidth}\raggedright
Model
\end{minipage} & \begin{minipage}[c]{\linewidth}\centering
Best init. value
\end{minipage} & \begin{minipage}[c]{\linewidth}\centering
Early nAUC (10--20)
\end{minipage} & \begin{minipage}[c]{\linewidth}\centering
nAUC-BO (10--40)
\end{minipage} & \begin{minipage}[c]{\linewidth}\centering
Late-stage nAUC (20--40)
\end{minipage} & \begin{minipage}[c]{\linewidth}\raggedright
Exact-optimum hits
\end{minipage} \\
\midrule
\noalign{}
\endfirsthead
\caption[]{\textbf{(continued)}}\\
\toprule
\noalign{}
\begin{minipage}[c]{\linewidth}\raggedright
Model
\end{minipage} & \begin{minipage}[c]{\linewidth}\centering
Best init. value
\end{minipage} & \begin{minipage}[c]{\linewidth}\centering
Early nAUC (10--20)
\end{minipage} & \begin{minipage}[c]{\linewidth}\centering
nAUC-BO (10--40)
\end{minipage} & \begin{minipage}[c]{\linewidth}\centering
Late-stage nAUC (20--40)
\end{minipage} & \begin{minipage}[c]{\linewidth}\raggedright
Exact-optimum hits
\end{minipage} \\
\midrule
\noalign{}
\endhead
\bottomrule
\noalign{}
\endlastfoot
DeepSeek 4.0 Flash & 16.128 ± 2.777 & 0.9266 & 0.9698 ± 0.0395 & 0.9914 ±
0.0151 & 5/5, latest at evaluation 28 \\
DeepSeek 4.0 Pro & 17.281 ± 1.860 & 0.956 & 0.9817 ± 0.0308 & 0.9946 ±
0.0083 & 5/5, latest at evaluation 26 \\
Qwen3-Max & 18.113 ± 0.000 & 0.9837 & 0.9910 ± 0.0063 & 0.9947 ± 0.0081
& 3/5, reached at evaluations 15--23 \\
GPT-5.6-sol & 18.113 ± 0.000 & 0.9849 & 0.9912 ± 0.0066 & 0.9944 ±
0.0070 & 5/5, latest at evaluation 39 \\
\end{longtable}
}

\section{Metrics and Experimental Configuration}\label{app:B}

\subsection{Metric Definitions}\label{app:B.1}

The objective value is returned by the table oracle, and the global
optimum is restricted to the given candidate table. The normalization
and discrete integration for \(Q_n\) are defined in Appendix B.2.1; the
threshold-reaching time is defined in Appendix B.2.2.

{
\begin{longtable}[]{@{}
  >{\raggedright\arraybackslash}p{(\linewidth - 6\tabcolsep) * \real{0.1500}}
  >{\raggedright\arraybackslash}p{(\linewidth - 6\tabcolsep) * \real{0.2000}}
  >{\centering\arraybackslash}p{(\linewidth - 6\tabcolsep) * \real{0.1300}}
  >{\raggedright\arraybackslash}p{(\linewidth - 6\tabcolsep) * \real{0.5200}}@{}}
\caption[Definitions and interpretation of the reported metrics]{\textbf{Definitions and interpretation of the reported metrics.}}\label{tab:s-metrics}\\
\toprule
\noalign{}
\begin{minipage}[c]{\linewidth}\raggedright
Metric
\end{minipage} & \begin{minipage}[c]{\linewidth}\raggedright
Definition
\end{minipage} & \begin{minipage}[c]{\linewidth}\centering
Direction
\end{minipage} & \begin{minipage}[c]{\linewidth}\raggedright
Interpretation
\end{minipage} \\
\midrule
\noalign{}
\endfirsthead
\caption[]{\textbf{(continued)}}\\
\toprule
\noalign{}
\begin{minipage}[c]{\linewidth}\raggedright
Metric
\end{minipage} & \begin{minipage}[c]{\linewidth}\raggedright
Definition
\end{minipage} & \begin{minipage}[c]{\linewidth}\centering
Direction
\end{minipage} & \begin{minipage}[c]{\linewidth}\raggedright
Interpretation
\end{minipage} \\
\midrule
\noalign{}
\endhead
\bottomrule
\noalign{}
\endlastfoot
Best-so-far & \(I_B\) & Task-dependent & The best observed
objective value at the end of the budget, retaining the original
physical units. \\
nAUC-BO & \(\frac{1}{B-10}\int_{10}^{B}Q_n\,dn\) & ↑ & The normalized
incumbent area over evaluations 10--40, measuring cumulative search
quality during the BO stage. \\
EF40 & \(Q_{40}\) & ↑ & Full-table min--max-normalized endpoint quality;
1 denotes the best table value. Exact candidate-ID hits are counted separately. \\
Late-stage nAUC & \(\frac{1}{40-20}\int_{20}^{40}Q_n\,dn\) & ↑ & The
normalized incumbent area over evaluations 20--40, describing cumulative
quality in the second half of the run. \\
\end{longtable}
}

\subsection{Experimental Configuration}\label{app:B.2}

\subsubsection{Normalization and Discrete Integration}\label{app:B.2.1}

\(y_\star\) and \(y_{\mathrm{worst}}\) are the best and worst objectives
over the complete candidate table for the task and are fixed before
search. For a maximization task, normalized quality is

\[
Q_n=\frac{I_n-y_{\mathrm{worst}}}{y_\star-y_{\mathrm{worst}}},
\]

and for a minimization task it is

\[
Q_n=\frac{y_{\mathrm{worst}}-I_n}{y_{\mathrm{worst}}-y_\star}.
\]

The discrete trapezoidal averages in Section 5.3 of the main text are computed on
\(n=10,\ldots,40\) as a trapezoidal mean: nAUC-BO averages over the
adjacent 30 segments, and late-stage nAUC averages over the adjacent 20
segments from \(n=20\) to \(40\). Full-table extrema are used for
offline normalization and are not included in the information available
to the optimizer in Section 5.4.

\subsubsection{Threshold Reaching and Exact Hits}\label{app:B.2.2}

\(t_{95}\) is, for each seed, the oracle evaluation index of the first
\(Q_n\ge0.95\), counted from evaluation 1 and including the 10
initialization evaluations. Seeds that do not reach the threshold within
the budget \(B=40\) are assigned the penalty value \(B+1=41\) and
included in the mean over all five seeds; the number of seeds reaching
the threshold is reported separately.

Exact optimum hits in the optimization experiments are counted using the
canonical best ID selected by the evaluation definition; other IDs tied
for the optimum are not counted separately for this metric. Exact hits
and 95\% threshold reaching are reported separately. The
zero-observation audit in Section 5.10 instead counts hits of the single
anchor ID specified for each repetition over the first 40 evaluations;
the three tied Suzuki optimum IDs are rotated as anchors.

\subsubsection{Result Sources and Record Scope}\label{app:B.2.3}

Aggregate values are taken from archived evaluation records and are available from the authors upon reasonable request. COF uses the 605-row finite candidate table frozen
by LABO {[}23{]}, whose objective is high-fidelity GCMC equilibrium
adsorption selectivity; the relevant scientific background is given in
{[}24{]}. The source study reports 609 structures, and a row-by-row
mapping to the 605-row table has not been established.

PCE10 uses the LABO finite table {[}23{]}.
The inputs are the composition fractions of PCE10, P3HT, PCBM, and
oIDTBR, and the target field is \texttt{degradation}; the table contains
1,040 records and 1,020 distinct compositions.

\subsubsection{Segmented Area Comparison across Models}\label{app:B.2.4}

Under the same normalization and integration rules, 30 times nAUC-BO
equals 10 times early nAUC plus 20 times late-stage nAUC. Decomposing
the reported means, approximately 82\%, 90\%, and 91\% of the BO-stage
area advantages of DeepSeek 4.0 Pro, Qwen3-Max, and GPT-5.6-sol over
DeepSeek 4.0 Flash, respectively, lie in the 10--20 interval.

\subsubsection{Baseline Sources, Registered Versions, and Adaptation Scope}\label{app:B.2.5}

The table below lists the registered baseline implementations and their
adaptations to finite tables. The first 12 characters of each commit
identifier are retained.

{
\begin{longtable}[]{@{}
  >{\raggedright\arraybackslash}p{(\linewidth - 4\tabcolsep) * \real{0.3333}}
  >{\raggedright\arraybackslash}p{(\linewidth - 4\tabcolsep) * \real{0.3333}}
  >{\raggedright\arraybackslash}p{(\linewidth - 4\tabcolsep) * \real{0.3333}}@{}}
\caption[Baseline implementation sources, registered versions, and adaptation scope]{\textbf{Baseline implementation sources, registered versions, and adaptation scope.}}\label{tab:s-implementations}\\
\toprule
\noalign{}
\begin{minipage}[c]{\linewidth}\raggedright
Source label and reference
\end{minipage} & \begin{minipage}[c]{\linewidth}\raggedright
Implementation source and registered version
\end{minipage} & \begin{minipage}[c]{\linewidth}\raggedright
Implementation and adaptation scope
\end{minipage} \\
\midrule
\noalign{}
\endfirsthead
\caption[]{\textbf{(continued)}}\\
\toprule
\noalign{}
\begin{minipage}[c]{\linewidth}\raggedright
Source label and reference
\end{minipage} & \begin{minipage}[c]{\linewidth}\raggedright
Implementation source and registered version
\end{minipage} & \begin{minipage}[c]{\linewidth}\raggedright
Implementation and adaptation scope
\end{minipage} \\
\midrule
\noalign{}
\endhead
\bottomrule
\noalign{}
\endlastfoot
Vanilla BO {[}3--5{]} & In-house implementation &
Numerical GP baseline \\
HEBO {[}29{]} &
\href{https://github.com/huawei-noah/HEBO}{huawei-noah/HEBO},
\texttt{ee6112d39d1a} & Finite-table adapter with unqueried-candidate mapping; quasi-random
fallback after repeated failure (see B.2.5) \\
LABO {[}23{]} &
\href{https://github.com/Xinzhe309/LABO}{Xinzhe309/LABO},
\texttt{c97d5615ebe8} & Low-fidelity prediction and candidate selection
through the upstream adapter \\
CAKE {[}19{]} &
\href{https://github.com/richardcsuwandi/cake}{richardcsuwandi/cake},
\texttt{ece5cbc04b3e} & Kernel search and finite-table selection through
the upstream adapter \\
LMABO {[}21{]} &
\href{https://github.com/giang-n-ngo/lmabo}{giang-n-ngo/lmabo},
\texttt{b1671f5ad79f} & Adapted acquisition selection; parse failures raise explicit errors
instead of silent UCB fallback \\
BORA {[}18{]} &
\href{https://github.com/Ablatif6c/bora-the-explorer}{Ablatif6c/bora-the-explorer},
\texttt{83b2ebf86f4f} & Region guidance and finite-table candidate
handling \\
Reasoning BO {[}30{]} &
\href{https://github.com/little1d/Reasoning-BO}{little1d/Reasoning-BO},
\texttt{4df3cd64abd1} & Finite-table rewrite (GP shortlist + LLM point
selection); bypasses the Ax/Camel workflow \\
LLAMBO {[}17{]} &
\href{https://github.com/tennisonliu/LLAMBO}{tennisonliu/LLAMBO},
\texttt{196fe237f60a} & Candidate mapping and independent sampling
expansion \\
\end{longtable}
}

The HEBO adapter uses the authors' \texttt{HEBO} class, constructs
numerical or categorical spaces according to the input type, and maps
suggestions to unqueried candidates in the table. If suggestion
generation fails, it calls \texttt{quasi\_sample}; after two consecutive
failures, this fallback is used for all remaining rounds. The returned
record contains the fallback count and whether fallback remained active.

\subsubsection{BH and Suzuki Data Sources and Candidate Tables}\label{app:B.2.6}

The BH data description cites Ahneman et al.~{[}31{]}, and the Suzuki
data description cites Perera et al.~{[}32{]}. Candidate tables
underwent column-name normalization and IUPAC-name mapping; their sizes
and processing are listed below.

{
\begin{longtable}[]{@{}
  >{\raggedright\arraybackslash}p{(\linewidth - 4\tabcolsep) * \real{0.1200}}
  >{\raggedright\arraybackslash}p{(\linewidth - 4\tabcolsep) * \real{0.3600}}
  >{\raggedright\arraybackslash}p{(\linewidth - 4\tabcolsep) * \real{0.5200}}@{}}
\caption[BH and Suzuki candidate tables and data processing]{\textbf{BH and Suzuki candidate tables and data processing.}}\label{tab:s-data}\\
\toprule
\noalign{}
\begin{minipage}[c]{\linewidth}\raggedright
Task
\end{minipage} & \begin{minipage}[c]{\linewidth}\raggedright
Rows and unique input combinations
\end{minipage} & \begin{minipage}[c]{\linewidth}\raggedright
Processing and protocol boundary
\end{minipage} \\
\midrule
\noalign{}
\endfirsthead
\caption[]{\textbf{(continued)}}\\
\toprule
\noalign{}
\begin{minipage}[c]{\linewidth}\raggedright
Task
\end{minipage} & \begin{minipage}[c]{\linewidth}\raggedright
Rows and unique input combinations
\end{minipage} & \begin{minipage}[c]{\linewidth}\raggedright
Processing and protocol boundary
\end{minipage} \\
\midrule
\noalign{}
\endhead
\bottomrule
\noalign{}
\endlastfoot
BH & 790 rows; 790 unique combinations & Fixed-product Buchwald\_sub4 background with four variable categorical fields; the benchmark excludes the 35 cross-product training priors \\
Suzuki & 5,760 rows; 5,760 unique combinations & Five categorical fields; the
29/5,731 train/test split in the data description is not equivalent to
the 10+30 query budget reported here \\
\end{longtable}
}

Input-feature checks found no duplicate combinations in either the BH or Suzuki table.

\section{Search Process and Implementation Audits}\label{app:C}

\subsection{Search-Process Cases}\label{app:C.1}

The cases below describe strategy changes and observation associations
within individual runs. They are process examples only and do not serve
as evidence for the main-text performance comparison or for mechanistic
causal claims. P3HT/CNT and OER are additional tasks and are not
included in the four-task benchmark comparison in the main text.

\begin{itemize}
\tightlist
\item
  \textbf{COF Xe/Kr separation: multi-source candidates and pore
  structure.} In the case record, seed 3 found the table optimum of
  approximately 18.48 at the third BO round; candidates came from both
  acquisition and max-variance, review changed their ranking, and the
  policy cited 34 historical evidence items. The run record proposed
  hypotheses that micropores below approximately 7 Å, a void fraction
  near 0.2, and N/metal features were associated with high-value
  regions; this run did not independently validate those hypotheses
  statistically. Most of the corresponding domain hypotheses remained
  marked INCONCLUSIVE.
\item
  \textbf{P3HT/CNT conductivity: retrieval of historical failures and
  strategy switching.} In the case record, seed 2 found 838.31 S/cm at
  the 28th BO round; it first retrieved failure records for
  SAAS-GP+LogEI and Deep Ensemble, then switched back to
  SAAS-GP+UCB(β=3)+Global. Two samples near 50\% D1 behaved differently:
  49.9\% corresponded to 286.1 S/cm, whereas 50.3\% corresponded to
  838.3 S/cm.
\item
  \textbf{PCE10 photostability: coverage candidates after a plateau.}
  Seed 3 found the smallest degradation value in the table,
  approximately 0.0016, through a farthest-distance candidate rather
  than the top acquisition candidate. The preceding four rounds were on
  a plateau, and the record indicates that the policy adjusted the
  candidate-source weights according to uncovered composition regions.
\item
  \textbf{Buchwald--Hartwig: controlled comparison and hypothesis
  revision.} Seed 0 found the table optimum of approximately 86.60\% at
  round 12. The preceding record included a controlled comparison with
  ligand, additive, and substrate fixed while only the base was changed;
  the yield varied from 81.57\% to 28.39\%. Another controlled
  comparison targeted the hypothesis that ligand effects under MTBD were
  smaller than 5 pp; the median effect was approximately 9.12 pp, and
  the record marked the hypothesis as REFUTE. This judgment applies only
  to that controlled comparison.
\item
  \textbf{Suzuki--Miyaura: strategy switching after a plateau.} After 21
  plateau rounds, the policy retrieved historical records for base
  conditions and successful early UCB(β=2) runs, then switched to
  SAAS-GP+UCB(β=2)+TuRBO and found 98.69\% using an exploitation-biased
  shortlist. The case also reported the yield ordering LiOtBu
  $>$ K3PO4 $>$ KOH $>$ NaHCO3 ≈ NaOH
  $>$ CsF under specific controlled conditions; the LiOtBu
  hypothesis was supported in two controlled groups, with a support rate
  of 1.0. This ordering applies to the controlled reaction conditions in
  the record.
\item
  \textbf{OER: retaining the formulation during the small-sample stage.}
  After pure Mn exploration produced an overpotential of approximately
  0.517 V, the policy kept GP+UCB for about 10 observations and
  subsequently found Ni0.6Fe0.3Ce0.1 with an overpotential of
  approximately 0.342 V. The Ni--Fe--La region proposed for exploration
  was not covered in the record.
\end{itemize}

Figure~\ref{fig:C1} covers 20 runs across four tasks. The effective surrogate
model switched at least once in every run and at most 12 times; the
acquisition function and search strategy switched less often, with 9
runs each remaining unchanged throughout. No \texttt{STOP} occurred in
these runs.

\begin{figure}[p]
\centering
\includegraphics[width=\linewidth,height=0.82\textheight,keepaspectratio]{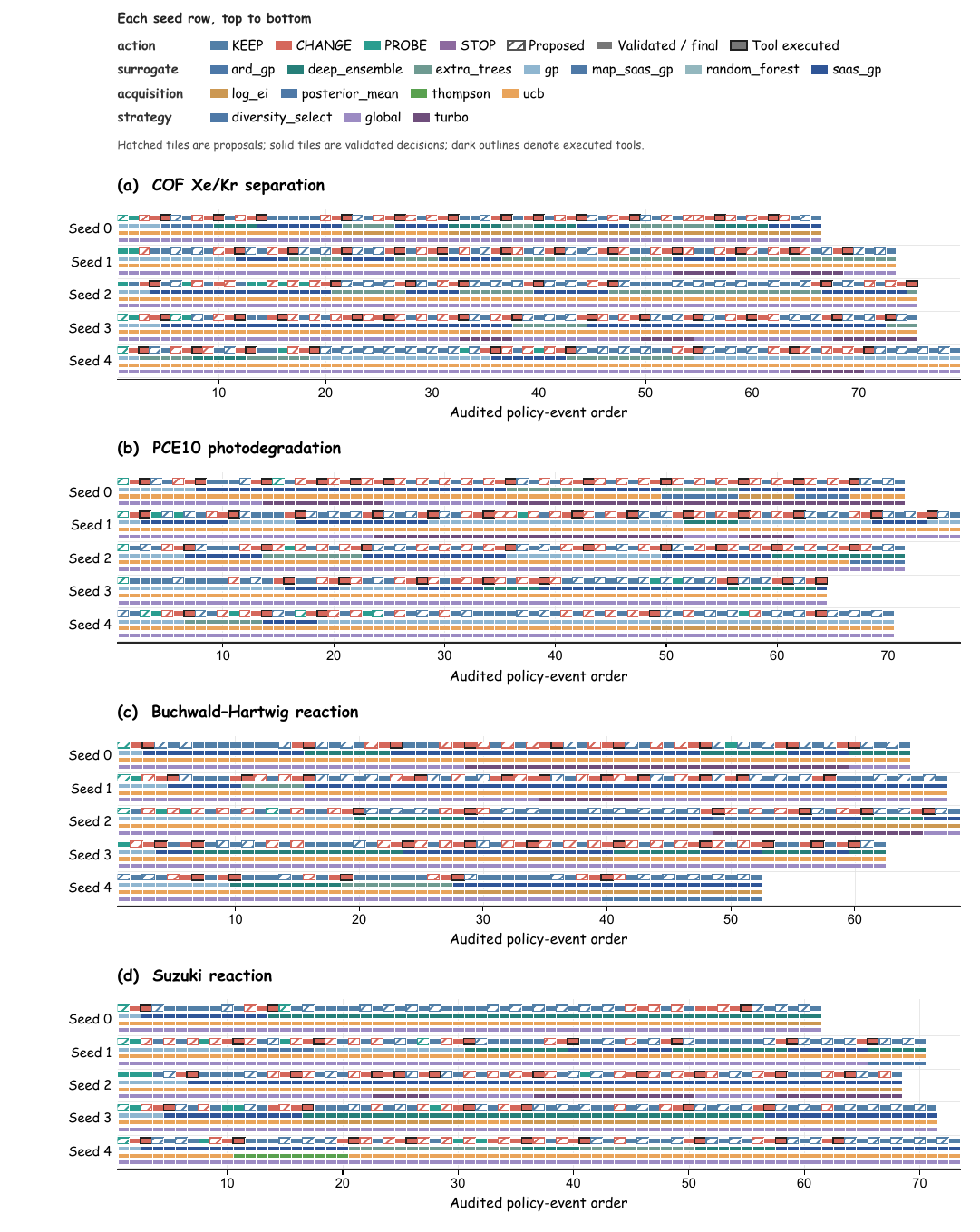}
\captionof{figure}{\textbf{Strategy events and effective formulations
across four tasks.} (a) COF; (b) PCE10; (c) BH; (d) Suzuki. Each row
corresponds to one seed. From top to bottom, the four bands represent
strategy actions, the effective surrogate model, the effective
acquisition function, and the effective search strategy. Hatching
denotes a strategy draft, solid fill denotes a validated strategy
proposal that was adopted, and a dark border denotes execution of an
algorithmic tool; the last three formulation bands record the formulation that
actually became effective after each event. The horizontal axis is the
temporal order of audit events and does not correspond to oracle
evaluation count.}\label{fig:C1}
\end{figure}

\subsection{Heuristic Branin Run Trajectory}\label{app:C.2}

A deterministic Branin run illustrates the recorded search process but is
not part of the benchmark. It used a heuristic strategy and did not invoke
an LLM. The settings were seed 42, three initial observations, and
a total budget of 6. The run completed three optimization rounds,
changed the acquisition function from UCB to logarithmic expected
improvement, then changed global search to TuRBO, and saved complete
records of strategies, candidate shortlists, reviews, selections,
observations, and memory updates.

The best value was already found during initialization, and the
following three observations did not improve it.

\subsection{Hypothesis States, Public Reflection, and Memory Rules}\label{app:C.3}

\subsubsection{Implementation Contract}\label{app:C.3.1}

{
\begin{longtable}[]{@{}
  >{\raggedright\arraybackslash}p{(\linewidth - 2\tabcolsep) * \real{0.5000}}
  >{\raggedright\arraybackslash}p{(\linewidth - 2\tabcolsep) * \real{0.5000}}@{}}
\caption[Hypothesis, reflection, and research-memory mechanisms]{\textbf{Hypothesis, reflection, and research-memory mechanisms.}}\label{tab:s-memory}\\
\toprule
\noalign{}
\begin{minipage}[c]{\linewidth}\raggedright
Object
\end{minipage} & \begin{minipage}[c]{\linewidth}\raggedright
Current mechanism
\end{minipage} \\
\midrule
\noalign{}
\endfirsthead
\caption[]{\textbf{(continued)}}\\
\toprule
\noalign{}
\begin{minipage}[c]{\linewidth}\raggedright
Object
\end{minipage} & \begin{minipage}[c]{\linewidth}\raggedright
Current mechanism
\end{minipage} \\
\midrule
\noalign{}
\endhead
\bottomrule
\noalign{}
\endlastfoot
Method hypothesis \texttt{METHOD} & States the claim, prediction, and
falsification condition; does not include a domain contrast \\
Domain main-effect and interaction hypotheses & Must declare
\texttt{domain\_contrast}, including comparison arms, controlled
factors, and a threshold; both types share the evaluator \\
Contrast evaluation & Uses observed records only; exact-control matching
by default (optional tolerance). Judgments derive from group counts,
thresholds, and direction \\
Per-hypothesis reflection & SUPPORT/REFUTE/INCONCLUSIVE, with evidence
and a public rationale; missing an explicit verdict that may be cited is
downgraded to uncertain; DIFFERENT propositions are handled under their
own threshold rule when sufficient groups exist \\
Hypothesis state & Stores supporting/opposing citations, evaluation
counts, and state changes, which feed the subsequent Policy \\
Playbook & Citations are resolved to experimental events in the current
run and deduplicated by round and candidate; it may be activated by
default when at least two supporting items exist and no resolved
experimental counterevidence exists; counterevidence in unresolved
citations is outside this judgment \\
TaskMemory & Component-condition statistics over a default recent
50-round window, which may feed review weights; reward mixes level,
improvement, and information proxy \\
\end{longtable}
}

Public reflection treats explicitly recorded hypotheses, evidence, and
evaluations as research state. Domain support is established by
observations that satisfy declared contrast conditions and minimum
evidence requirements; unmatched observations are not counted as
support. Playbook rules are retained or downgraded when evidence is
insufficient or counterevidence appears; the rules do not alter the
execution contract.

\subsubsection{Citation Fields and Validation Scope}\label{app:C.3.2}

\texttt{profile\_evidence} is normalized to a list of strings and
checked for non-emptiness along the corresponding paths; this check does
not verify source existence or visibility. The top-level
\texttt{memory\_evidence\_ids} is checked on some controller paths for
membership in the visible set; existence filtering also depends on the
Registry supplied at call time. Candidate-plan \texttt{evidence\_ids} do
not inherit the top-level validation. Visibility filtering for
reflection citations and domain-contrast judgment are performed
separately, and their validation scopes differ by path.

The mechanical three-state result is computed from task facts, state,
and diagnostics; historical notes influence subsequent state through
policy selection and do not directly modify \texttt{selection\_gate}.
Disabling \texttt{research.enabled} still retains TaskMemory,
review-history feedback, and trajectory-component summaries.

\subsubsection{Evaluation Scope of Cross-Round Feedback}\label{app:C.3.3}

Hypothesis-evaluation and state-change fields are not populated
consistently in the available reflection logs. Consequently, these logs
do not support reconstruction of the full sequence linking hypotheses,
observations, revisions, and subsequent actions. Appendix C.2 provides an
example of the event records captured in a deterministic heuristic run.

\subsubsection{Optional Proposal Review and Candidate Review}\label{app:C.3.4}

The optional proposal audit in Fig. 1 corresponds to conditional review
in the structured strategy controller. The flow is requested only when
research discussion is enabled and either a stagnation state occurs or
the diagnostic utility of an alternative surrogate exceeds that of the
current surrogate and a configured threshold. The method, domain, and
counterevidence critics review the same policy draft and return SUPPORT,
REVISE, or BLOCK recommendations; the arbiter aggregates them into
ACCEPT or REVISE, with at most one proposal revision triggered by the
latter. The final proposal still passes the ordinary runtime guards.

A critic's BLOCK is a proposal-review opinion and has no authority to
hard-veto a candidate or submit an experiment. If the review service or
context preparation fails, the flow may return a single policy draft
handled by the runtime guards. Historical retrieval precedes final
action review; memory retrieval and diagnostic PROBE use independent
round quotas.

Candidate review returns task-dimension scores for the shortlist formed
by the execution layer, for fusion in Section 4.5; post-observation
reflection instead links actual results to public hypotheses. Proposal
review, candidate review, and reflection concern policy proposals,
candidate sets, and observed evidence, respectively. Candidate fusion
uses valid review dimensions that cover the complete shortlist and
weights them by dimension; reviewer confidence is not used in the
calculation. Online, if no complete valid dimension with a positive
total weight is available, selection falls back to purely numerical
choice; other anomalies follow their respective error paths. Warm start
executes the initial experiment only after obtaining at least one
complete review dimension.

\subsection{Declaration Protocol and Current Implementation}\label{app:C.4}

{
\begin{longtable}[]{@{}
  >{\raggedright\arraybackslash}p{(\linewidth - 4\tabcolsep) * \real{0.3333}}
  >{\raggedright\arraybackslash}p{(\linewidth - 4\tabcolsep) * \real{0.3333}}
  >{\raggedright\arraybackslash}p{(\linewidth - 4\tabcolsep) * \real{0.3333}}@{}}
\caption[Declared protocol and current implementation]{\textbf{Declared protocol and current implementation.}}\label{tab:s-protocol}\\
\toprule
\noalign{}
\begin{minipage}[c]{\linewidth}\raggedright
Stage
\end{minipage} & \begin{minipage}[c]{\linewidth}\raggedright
Declaration protocol
\end{minipage} & \begin{minipage}[c]{\linewidth}\raggedright
Current implementation
\end{minipage} \\
\midrule
\noalign{}
\endfirsthead
\caption[]{\textbf{(continued)}}\\
\toprule
\noalign{}
\begin{minipage}[c]{\linewidth}\raggedright
Stage
\end{minipage} & \begin{minipage}[c]{\linewidth}\raggedright
Declaration protocol
\end{minipage} & \begin{minipage}[c]{\linewidth}\raggedright
Current implementation
\end{minipage} \\
\midrule
\noalign{}
\endhead
\bottomrule
\noalign{}
\endlastfoot
Atomic admission & The complete combination satisfies the conditions for
the current round & Primarily checks atomic changes in CHANGE and
retains atoms that can continue execution \\
Supporting material & The required support list is non-empty & Verifies list non-emptiness only; citation validity and premise
entailment unverified \\
Portfolio & Each acquisition capability and the final surrogate are
compatible and satisfy admission & Filters BLOCKED IDs and verifies interfaces; partial binding for
NEEDS\_EVIDENCE; zero-quota keys still checked \\
CHANGE→KEEP & Recheck the candidate plan under the final formulation &
Candidate plan validated pre-guard; deferred components bypass
re-validation \\
Candidate and fusion & Bounded candidates, valid reviews, deterministic
submission & Decoupled quotas/slots; falls back to numerical utility if valid
review dimensions are absent \\
Research feedback & Public state is available for subsequent retrieval &
Typed hypotheses, reflections, and retrieval rules are saved \\
Budget and objective & Only oracle observations are appended & Ordinary
single-point paths separate query computation from objective
computation \\
Path coverage & The same checks apply to non-\texttt{STOP}
candidate-submission paths & The heuristic path does not cover all
checks applied by the main path \\
Termination and failure & \texttt{STOP} returns a report; no candidate
means no evaluation & Immediate report generation on \texttt{STOP}; empty candidate set
raises selection failure \\
Parameter state & Active parameters agree with actual execution & The
shared parameter dictionary may retain inactive fields \\
ID--feature consistency & Validation supports an adversarial consistency
check & Checks basic ID--feature mappings; adversarial consistency
unverified \\
\end{longtable}
}

\section{Figure Notes and Projection Examples}\label{app:D}

\subsection{Figure Notes}\label{app:D.1}

Main-text Figure 1 organizes the workflow around strategy adjustment, candidate
execution and observation, and research-memory feedback; the partitions
indicate functional order, while execution permissions are specified by
the individual roles. Main-text Figure 2 expands the capability contract,
evidence-condition compilation, two-level typed proposals, and execution
submission; paired E markers indicate that computational evidence is
returned to the compiler input. Both figures are mechanism schematics;
the scope of checks in the current implementation is given in Section
4.8 of the main text and Appendix C.4.

HEBO and Reasoning BO correspond to references {[}29--30{]} in the main
text, and the reaction data
correspond to references {[}31--32{]}.

\subsection{Two-Dimensional Projection Examples of Search Paths}\label{app:D.2}

Figure~\ref{fig:D1} shows the positions and order of incumbent updates in one run
for each of the four tasks, providing a process example for the
performance curves in the main text. The backgrounds are post hoc
explanatory projections and were not provided as inputs to the
optimizer.

\begin{figure}[htbp]
\centering
\includegraphics[width=\linewidth]{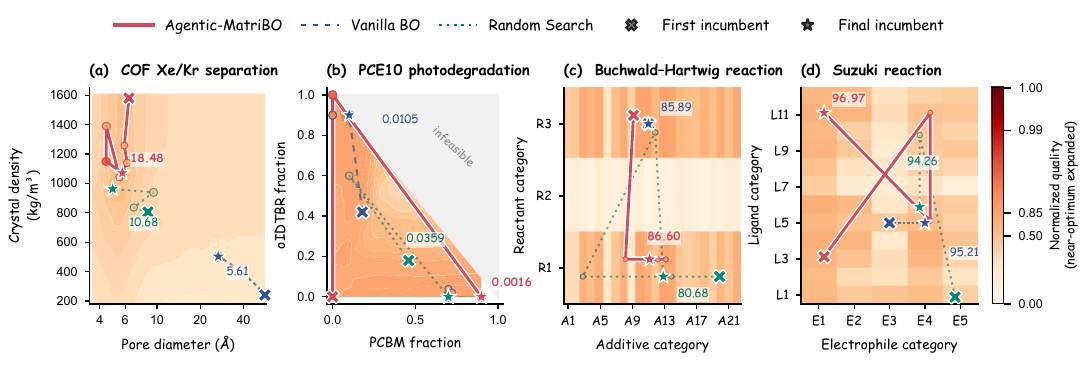}
\caption{\textbf{Two-dimensional projections and
incumbent-update examples.} (a) COF; (b) PCE10; (c) BH; (d) Suzuki. The
red solid, blue dashed, and teal dotted lines correspond to the proposed
method, Vanilla BO, and Random Search, respectively; crosses and stars
mark the first and final incumbents in the displayed run. PCE10 is a
minimization objective, and the background shows direction-normalized
quality; the gray region indicates that the sum of the two plotted
components exceeds 1. The BH and Suzuki coordinates are categorical
labels, and adjacency does not represent continuous physical distance;
connecting lines indicate update order only.}\label{fig:D1}
\end{figure}

\endgroup

\end{document}